\documentclass[sigconf]{acmart}

\usepackage{subfigure}
\usepackage{multirow}
\usepackage[linewidth=0.5pt]{mdframed}
\usepackage{graphicx}
\usepackage[export]{adjustbox}

\usepackage{algorithmic}
\usepackage[ruled,linesnumbered]{algorithm2e}
\usepackage{bbm}
\usepackage{colortbl}
\usepackage{enumitem}
\usepackage{makecell}
\usepackage{hyperref}

\newtheorem{theorem}{theorem}[section]             
\newtheorem{property}[theorem]{\textbf{property}}           
\newtheorem{proposition}[theorem]{\textbf{proposition}}     
\newtheorem{lemma}[theorem]{lemma}                 

\definecolor{darkred}{RGB}{162, 0, 0}

\definecolor{darkblue}{RGB}{4, 6, 173}

\AtBeginDocument{%
  \providecommand\BibTeX{{%
    \normalfont B\kern-0.5em{\scshape i\kern-0.25em b}\kern-0.8em\TeX}}}

\setcopyright{acmcopyright}
\copyrightyear{2018}
\acmYear{2018}
\acmDOI{XXXXXXX.XXXXXXX}

\acmConference[Conference acronym 'XX]{Make sure to enter the correct
  conference title from your rights confirmation emai}{June 03--05,
  2018}{Woodstock, NY}
\acmPrice{15.00}
\acmISBN{978-1-4503-XXXX-X/18/06}

\begin{document}

\title{NodeReg: Mitigating the Imbalance and Distribution Shift Effects in Semi-Supervised Node Classification via Norm Consistency}


\author{Shenzhi Yang}
\affiliation{%
  \institution{Soochow University}
  \city{Suzhou}
  \country{China}}
\email{20225227031@stu.suda.edu.cn}

\author{Jun Xia}
\affiliation{%
  \institution{Westlake University}
  \city{Hangzhou}
  \country{China}}
\email{xiajun@westlake.edu.cn}

\author{Jingbo Zhou}
\affiliation{%
  \institution{Westlake University}
  \city{Hangzhou}
  \country{China}}
\email{zhoujingbo@westlake.edu.cn}

\author{Xingkai Yao}
\affiliation{%
  \institution{Soochow University}
  \city{Suzhou}
  \country{China}}
\email{20234227040@stu.suda.edu.cn}

\author{Xiaofang Zhang}
\affiliation{%
  \institution{Soochow University}
  \city{Suzhou}
  \country{China}}
\email{xfzhang@suda.edu.cn}







\renewcommand{\shortauthors}{Trovato and Tobin, et al.}

\begin{abstract}\label{sec-abstract}

Aggregating information from neighboring nodes benefits graph neural networks (GNNs) in semi-supervised node classification tasks. Nevertheless, this mechanism also renders nodes susceptible to the influence of their neighbors. For instance, this will occur when the neighboring nodes are imbalanced, or the neighboring nodes contain noise, which can even affect the GNN's ability to generalize out of distribution. We find that ensuring the consistency of the norm for node representations can significantly reduce the impact of these two issues on GNNs. To this end, we propose a regularized optimization method called \textbf{NodeReg}\footnote{NodeReg: \url{https://anonymous.4open.science/r/NodeReg-0FA1}} 
that enforces the consistency of node representation norms. This method is simple but effective and satisfies Lipschitz continuity, thus facilitating stable optimization and significantly improving semi-supervised node classification performance under the above two scenarios. To illustrate, in the imbalance scenario, when training a GCN with an imbalance ratio of 0.1, NodeReg outperforms the most competitive baselines by 1.4\%-25.9\% in F1 score across five public datasets. Similarly, in the distribution shift scenario, NodeReg outperforms the most competitive baseline by 1.4\%-3.1\% in accuracy.
\end{abstract}




\received{20 February 2007}
\received[revised]{12 March 2009}
\received[accepted]{5 June 2009}

\maketitle

\section{Introduction}\label{Sec-Introduction}

\begin{figure}[t]
	\centering
   \subfigure[Training w/o Norm Consistency]{
		\includegraphics[width=0.4\linewidth]{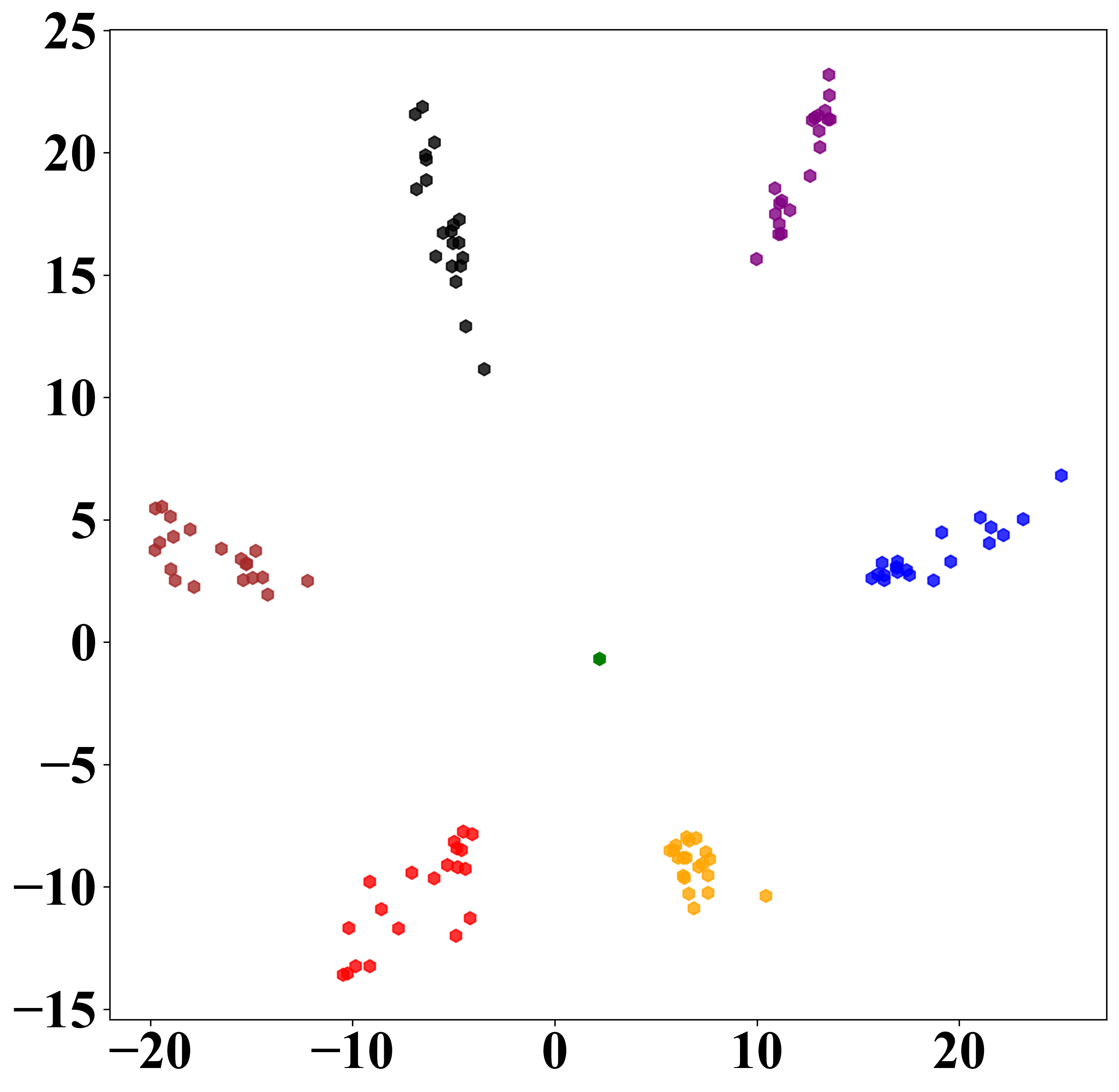}
 }
  \subfigure[Testing w/o Norm Consistency]{
		\includegraphics[width=0.4\linewidth]{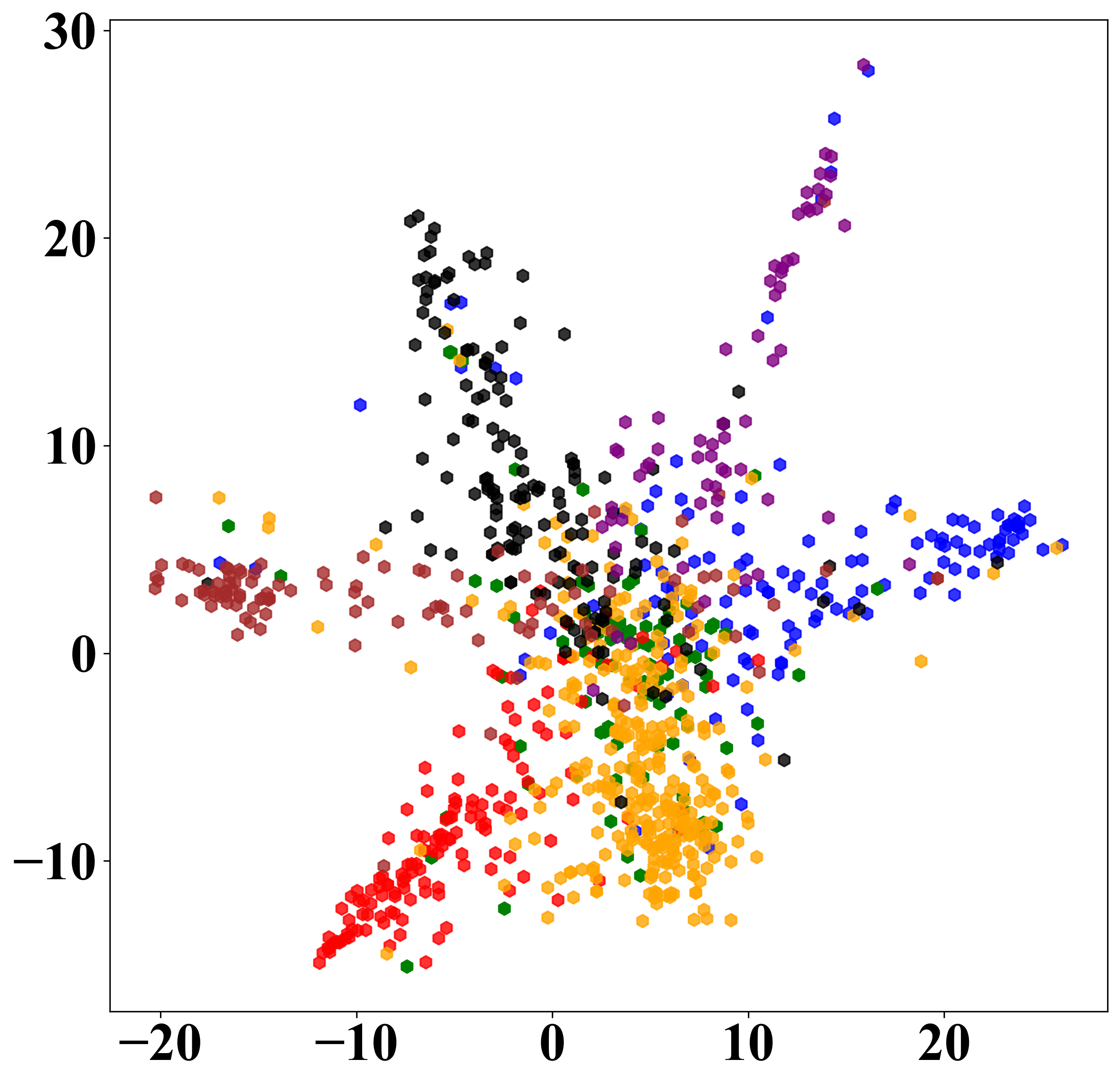}
 }

   \subfigure[Training w/ Norm Consistency]{
		\includegraphics[width=0.4\linewidth]{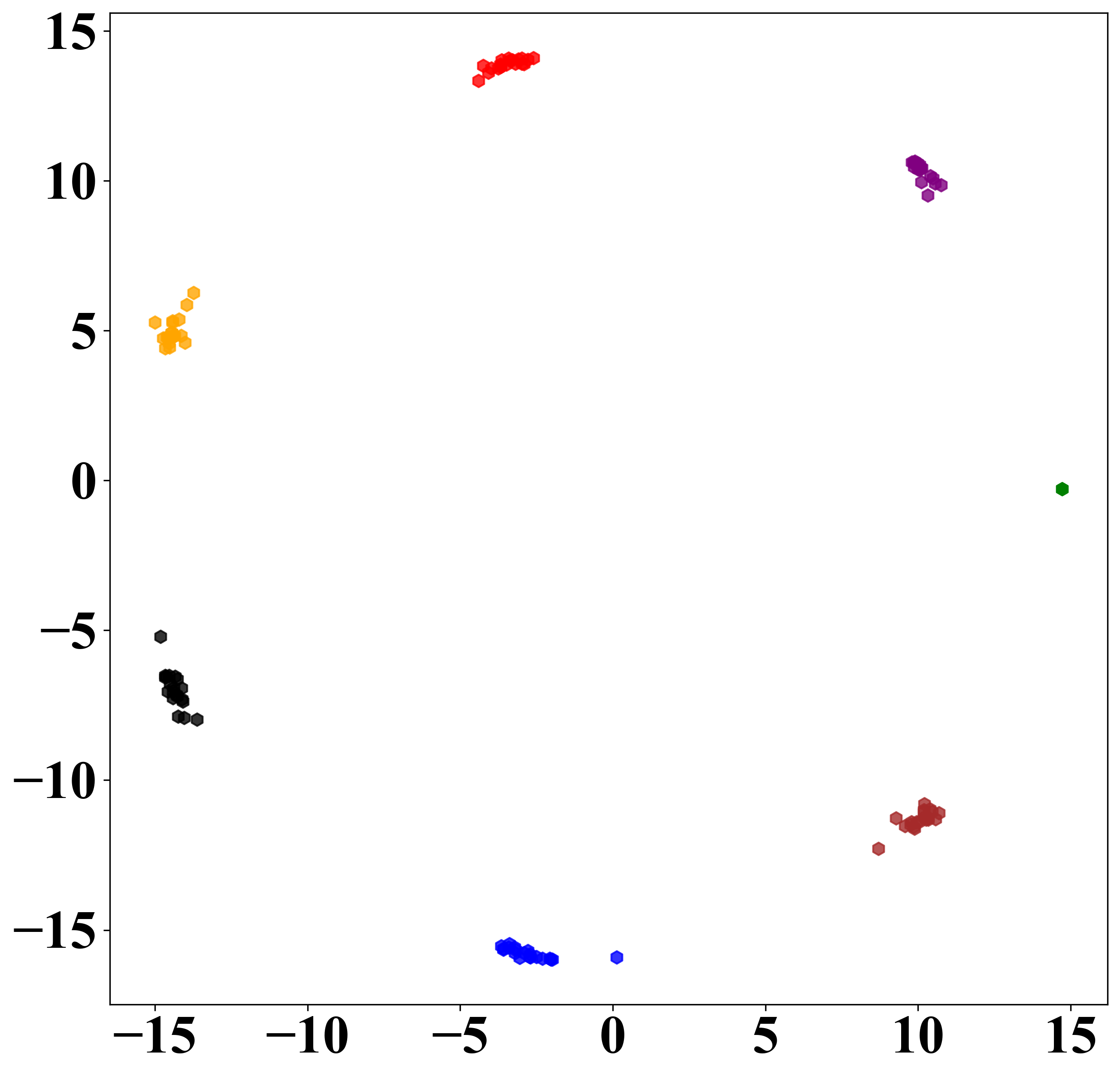}
 }
    \subfigure[Testing w/ Norm Consistency]{
		\includegraphics[width=0.4\linewidth]{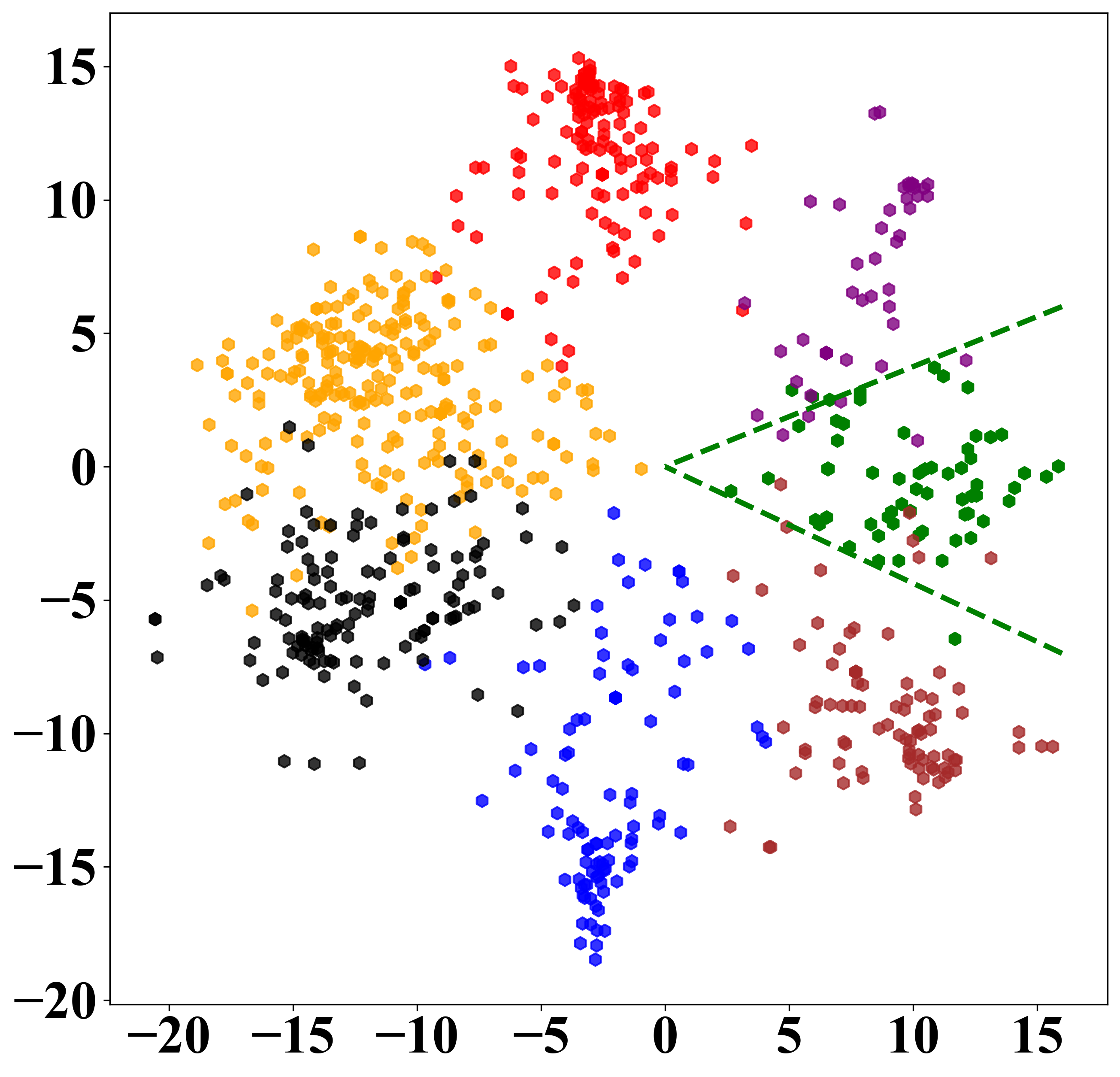}
 }
 \Description{Visualizing representations of nodes using GCN for node classification on the \textit{Cora} dataset. For the green class, we use only a single node for classification training to study the effect of consistent node representation norms under a node imbalance setting. The left-side figures (a) and (c) show the visualization of the trained node representations, while the right-side figures (b) and (d) visualize the representations of nodes to be predicted corresponding to (a) and (c), respectively.}
    \caption{Visualizing representations of nodes using GCN for node classification on the \textit{Cora} dataset. For the green class, we use only a single node for classification training to study the effect of consistent node representation norms under a node imbalance setting. The left-side figures (a) and (c) show the trained node representations, while the right-side figures (b) and (d) visualize the representations of nodes to be predicted corresponding to (a) and (c), respectively.}
    \label{F-Motivation}
\end{figure}

Graph neural networks (GNNs) \citep{kipf2016semi, 2017Graph, hamilton2017inductive} benefit from a unique information aggregation mechanism, which makes it very effective in processing data with graph structure. This has led to various applications in handling Internet-related tasks, such as Internet traffic and network routing optimization\citep{zhang2019inductive}, recommender systems\citep{fan2019graph}, knowledge graphs\citep{baek2020learning}, and information retrieval\citep{wang2019knowledge}.
In graph analysis, semi-supervised node classification is one of the most common tasks. It involves classifying nodes using a small number of labeled nodes and many unlabeled node representations. 
Despite GNNs performing well in semi-supervised node classification tasks, they rely on the assumption of a balanced distribution of node labels across different categories, which is challenging because annotation is resource-intensive. When this assumption is not met, the performance of GNNs to classify nodes can significantly degrade. The challenge of maintaining good classification performance of GNNs under conditions of imbalanced node categories is referred to as the node imbalance problem\citep{shi2020multi, chen2021topology,zhao2021graphsmote,park2021graphens,yan2024rethinking,zhang2024bim}.

Furthermore, when there is a shift in the distribution of nodes, it can also impact the representation of neighboring nodes, particularly when the neighbors include nodes that need to be predicted. We consider node distribution shift another imbalance problem, as it disrupts the balance between the trained model parameters and the node distribution. Distribution shifts can cause GNNs to make inaccurate predictions about node categories. The challenge of maintaining good classification performance of GNNs in the presence of node distribution shifts is referred to as the out-of-distribution (OOD) generalization problem\citep{arjovsky2019invariant,sun2016deep,ganin2016domain,sagawa2019distributionally,zhu2021shift,wu2022handling,wu2024graph} 
in semi-supervised node classification tasks.

Several methods have been proposed to tackle node imbalance. DR-GCN \citep{shi2020multi} uses specialized regularization layers to manage imbalanced data. In contrast, ReNode \citep{chen2021topology} introduces a metric to measure graph topology imbalance and provides a model-agnostic solution to address it. GraphSMOTE \citep{zhao2021graphsmote} generates pseudo nodes to rebalance the graph, and GraphENS \citep{park2021graphens} synthesizes the entire ego network to prevent overfitting to minor class neighbors. ReVar \citep{yan2024rethinking} reduces imbalance and model variance with regularization terms, and BIM \citep{zhang2024bim} highlights the influence of both class sample imbalance and receptive field imbalance between nodes. \textit{However, none of these approaches note that in node imbalance scenarios, the minority class tends to have a smaller norm.} 
For instance, we conduct experiments under an extremely imbalanced node setting. We visualize both the training set node representations and the test set node representations. Figure \ref{F-Motivation}(a) shows that, after training, the representation norms of minority class nodes are much smaller than those of majority class nodes, which 
leads to a blurred decision boundary and inaccurate predictions, as shown in Figure \ref{F-Motivation}(b). 

To tackle OOD generalization, several methods proposed like IRM \citep{arjovsky2019invariant}, DeepCoral\citep{sun2016deep}, DANN\citep{ganin2016domain}, and GroupDRO\citep{sagawa2019distributionally} offer robust learning algorithms to maintain stable predictions despite distribution shifts. Mixup \citep{zhang2017mixup} increases training data diversity through interpolation, while SR-GNN\citep{zhu2021shift} adjusts GNN models to align with distributional shifts between training and inference. EERM\citep{wu2022handling} employs adversarial training to explore environmental invariances, and CaNet\citep{wu2024graph} enhances GNNs' OOD generalization by preserving causal invariance among nodes. \textit{However, none of these methods note that among the nodes to be predicted, the node representations with smaller norms form a region of significant overlap with majority classes as shown in Figure \ref{F-Motivation}(b).} We believe that the node representations in this region are susceptible to distribution shifts that can produce incorrect classification results.   

 As a result of the above analysis, we have discovered an intrinsic link between the node imbalance and OOD generalization problems on graphs, i.e., the inconsistency of the node representation norm leads to the deterioration of GNNs' performance in both scenarios.
 Hence, we propose a simple but effective regularized optimization method that enforces the consistency of node representation norms, called \textbf{NodeReg}, which satisfies both Lipschitz continuity and smoothness, thus facilitating stable optimization and bringing significant improvements in semi-supervised node classification performance for GNNs under the above two scenarios simultaneously. 
For instance, when we maintain consistent representation norms between minority and majority class nodes during training, as illustrated in Figure \ref{F-Motivation}(c), the representation of minority class nodes to be predicted preserves a clear decision boundary, as shown in Figure \ref{F-Motivation}(d). This helps mitigate the node imbalance problem.
For the OOD generalization problem, by comparing Figure \ref{F-Motivation}(b) and Figure \ref{F-Motivation}(d), we observe that 
the representations of nodes to be predicted under the constraint of consistent norms form a ring-like distribution rather than a star-shaped one, reducing the overlap in the small norm region and enhancing intra-class cohesion. \textit{Since nodes are not independent, so larger norm representations can directly influence smaller norm representations during message propagation, the impact of node norms on semi-supervised node classification tasks is an urgent issue that warrants further research.}

Our key contributions can be summarized as follows:
\begin{itemize}[nosep, topsep=0pt, leftmargin=*]

\item \textbf{New Discovery}: We find for the first time an intrinsic link between the node imbalance and the OOD generalization problem in semi-supervised node classification tasks, i.e., the norm imbalance of node representations leads to the deterioration of the performance of GNNs in both scenarios.

\item \textbf{New Method}: Based on the discovery, we propose a simple but effective method, NodeReg, that enforces consistent node representation norms and has desirable optimization properties.

\item \textbf{Empirical Study}: We have conducted extensive experiments to show that NodeReg can significantly enhance the performance of GNNs in two scenarios: node imbalance and OOD generalization in semi-supervised node classification tasks. 

\item \textbf{Theoretical Analysis}: We provide a sound theoretical analysis of the effectiveness of our method from perspectives such as benign overfitting\citep{bartlett2020benign,cao2022benign,huang2023graph} and neural collapse\citep{papyan2020prevalence,zhu2021geometric,zhou2022all}.
\end{itemize}

\section{Related Works} \label{Sec-RelatedWork}
This section describes some works related to our research. Section \ref{subsec-GNN} introduces the basic paradigm of GNN. Section \ref{subSec-Imbalanced} presents related works on overcoming the node imbalance problem. Section \ref{subSec-OOD} discusses related works on addressing the OOD generalization issue. Section \ref{subSec-NCsis} introduces other relevant works that aim to achieve consistency of representation norms to varying degrees.
\subsection{Graph Neural Network}\label{subsec-GNN}
Let $\mathcal{G}=\{\mathcal{V}, \mathcal{E}, \boldsymbol{X}\}$ denote a graph, where $\mathcal{V}$, $\mathcal{E}$ and ${\boldsymbol X}_v \in \mathbb {R}^{|\mathcal{V}|\times d}$ denote the sets of nodes, edges,  and the matrix of features of the nodes, respectively. For the representation $\mathbf{h}_v$ of node $v$, its propagation of the $k$-th layer GNN is represented as:
\begin{equation}\label{equa-gnns}
   \mathbf{h}_v^{(k)} =\mathrm{UDT}^{(k)}\big(\mathrm{AGR}^{(k)}\big( \mathbf{h}_u^{(k-1)}:\forall u \in \mathcal{N}(v)\cup v \big) \big) 
\end{equation}
where $\mathrm{AGR}(\cdot)$ denotes GNNs aggregate representations from the node $v$ with its neighbors $\mathcal{N}(v)$ and $\mathrm{UDT}(\cdot)$ denotes GNNs update the representation of node $v$ with trainable parameters. The classic GNNs include GCN\citep{kipf2016semi}, GAT\citep{2017Graph}, GraphSAGE\citep{hamilton2017inductive} and etc.

We divide the set of nodes $\mathcal{V}$ into a training set $\mathcal{V}_{train}$  (which includes a set of labeled nodes $\mathcal{V}_l$ with label set $\boldsymbol{Y} = \{y_1, y_2,...,y_{|\mathcal{V}_l|} \}$ which is the label matrix for $c$ classes and a set of unlabeled nodes $\mathcal{V}_{u}$), a validation set $\mathcal{V}_{val}$, and a test set $\mathcal{V}_{test}$.
The optimization objective of semi-supervised node classification is:
\begin{equation}\label{equa-L-ce}
    \mathcal{L}_{CE} = \mathbb{E}_{v\sim \mathcal{V}_l} \big(- \sum_{i=0}^{c-1} y_i \log(\hat{y}_i) \big)
\end{equation}
where $\mathcal{L}_{CE}$ refers to the cross-entropy loss function and \( \hat{y}_i \) is the predicted probability generated by softmax function, i.e. $
  \hat{y}_i = \frac{e^{\mathbf{z}_i}}{\sum_{j=0}^{c-1} e^{\mathbf{z}_j}}
$ in which $\mathbf{z}$ denotes the logit.

\subsection{Imbalanced Node Classification}\label{subSec-Imbalanced}
The node imbalance classification problem refers to the situation where there is a significant discrepancy in the number of nodes from different classes within the $\mathcal{V}_{l}$, leading to poor classification performance for minority class nodes in the $\mathcal{V}_{u}$.

Recently, several methods have been proposed to address the issue of node imbalance. For example,
DR-GCN\citep{shi2020multi} uses a conditional adversarial regularization layer and a latent distribution alignment regularization layer to handle imbalanced data. ReNode\citep{chen2021topology} introduces a metric called Totoro, based on impact conflict detection, to measure the extent of graph topology imbalance and proposes a model-agnostic method. Inspired by the classical SMOTE algorithm\citep{chawla2002smote}, GraphSMOTE\citep{zhao2021graphsmote} rebalances nodes and edges by generating additional pseudo nodes. GraphENS\citep{park2021graphens} argues that GNNs often severely overfit the neighbors of minor class nodes and proposes a method that produces nodes by synthesizing the entire ego network. ReVar\citep{yan2024rethinking} mitigates the impact of imbalance and model variance by adding regularization terms across different views. Additionally, BIM\citep{zhang2024bim} suggests that node imbalance is influenced by class sample imbalance and an imbalance in the receptive field of nodes. Specifically, the imbalance of receptive fields refers to the differing number of K-hop neighbors across nodes, resulting in an uneven number of other nodes that a single node can influence. 
However, none of these approaches note that in node imbalance scenarios, the minority class tends to have a smaller norm, leading to less obvious decision boundaries for the minority class. 


\subsection{OOD Generalization on Graphs}\label{subSec-OOD}
GNNs perform excellently in semi-supervised node classification on in-distribution data, i.e., test nodes generated from the same distribution as the training data. However, when the test data distribution shifts, GNNs often perform unsatisfactorily on OOD nodes.

Invariant risk minimization (IRM)\citep{arjovsky2019invariant} is a learning paradigm to estimate invariant correlations across multiple training distributions. 
Deep Coral\citep{sun2016deep} extends Coral\citep{sun2016return} to learn a nonlinear transformation that aligns the correlation of layer activations. 
DANN\citep{ganin2016domain} introduces a new method for learning domain-adapted representations, in which the data at training and testing come from similar but different distributions. 
GroupDRO\citep{sagawa2019distributionally} shows that regularization is essential for worst group generalization in over-parameterized regimes, even if it is not for average generalization. By combining the group DRO model\citep{ben2013robust,namkoong2016stochastic} with increased regularization, GroupDRO achieves higher worst group accuracy.
 Mixup\citep{zhang2017mixup} aims to increase training data by interpolating between observed samples.  To explain the distributional discrepancy between the biased training data and the accurate inference distribution over graphs, SR-GNN\citep{zhu2021shift} is designed by adjusting the GNN model to account for the distributional shift between labeled nodes in the training set and the rest of the dataset. EERM\citep{wu2022handling} utilizes the invariance principle to develop an adversarial training method for environmental exploration. Recent work on CaNet\citep{wu2024graph} enhances the OOD generalization capability of GNNs by preserving causal invariance among nodes. Unlike these approaches, we reveal that norm imbalance impairs the OOD generalization ability of GNNs.


\subsection{Different Methods for Norm Consistency}\label{subSec-NCsis}
Some existing methods can make representation norms more consistent. For example, supervised contrastive learning loss $\mathcal{L}_{scl}$ \citep{Khosla_NIPS20_SupCon} promotes consistency of norms by pulling together representations of similar class samples and pushing apart those of different classes. Center loss $\mathcal{L}_{center}$ \citep{wen2016discriminative} achieves norm consistency among similar representations by iteratively updating class centers and minimizing the distance between samples of the same class and their respective class centers. However, these methods differ in several ways: they require label supervision, cannot achieve norm consistency between classes, and are inefficient. The $\mathcal{L}_{\mathrm{bound}}$ in NODESAFE \citep{yangbounded} directly aims to make node representation norms more consistent by minimizing the variance-to-mean ratio of the norms. However, it does not satisfy Lipschitz continuity, making optimization unstable, especially in imbalanced scenarios. In the theoretical analysis section \ref{subSec-method-other-NCsis} and the experimental section \ref{subSec-experiment-compare-other-NCsis}, we will compare these methods with our approach in detail.


\section{Method} \label{Sec-Method}
This section will introduce our method and conduct a theoretical analysis. Section \ref{subSec-NodeReg} provides a detailed explanation of our process. In Section \ref{subSec-Theory-analysis}, we analyze the favorable properties of our method. In Section \ref{subSec-method-other-NCsis}, we compare our method in detail with others that promote consistency in node representation norms to varying degrees, highlighting the differences and advantages of our approach.

\subsection{NodeReg: Enforcing the Consistency of Node Representation Norms} \label{subSec-NodeReg}


Based on the above analysis and observations of node representations (Figure \ref{F-Motivation}), we propose a simple but effective regularization optimization objective to achieve consistent norms for node representations.
First, we compute the mean $\bar{\mathcal{F}}$ of the Frobenius norm $\lVert \cdot \rVert_\mathcal{F}$ of the node representations (or logits) $\mathbf{z}$ as follows:
\begin{equation}\label{equa-NodeReg-mean}
    \bar{\mathcal{F}} = \text{stop-gradient} \big(\  \frac{1}{|\mathcal{V}|} \sum_{v \in \mathcal{V}} \lVert \mathbf{z}_v \rVert_\mathcal{F} \  \big)
\end{equation}
where $\lVert \mathbf{z}_v \rVert_\mathcal{F} = \sqrt{\sum\limits_{j=0}^{c-1} \mathrm{z}^2_j}$ which is equivalent to the $L$2 norm $\lVert \cdot \rVert_2$ here. The $\bar{\mathcal{F}}$ does not participate in gradient propagation.
  We then compute the difference between the ratio of each node's representation norm to the mean $\delta_v$ and 1, as follows:
\begin{equation}\label{equa-NodeReg-delta}
    \delta_v = 1 - \lVert \mathbf{z}_v \rVert_\mathcal{F} \cdot \bar{\mathcal{F}}^{-1}
\end{equation}
Our optimization goal is to drive $\delta_v \in (-\infty, 1)$ to converge to 0. We then apply further smooth-$L$1\citep{girshick2015fast} transformations to $\delta_v$ to make it more amenable to optimization, as follows:
\begin{equation}\label{equa-NodeReg-smoothl1}
    \mathcal{L}_{\mathrm{NodeReg}}^v = 
    \begin{cases} 
        \frac{1}{2\gamma} \cdot \delta^2_v & \text{if } |\delta_v| < \gamma, \\
        |\delta_v|-\frac{\gamma}{2} & \text{if } \delta_v \geq \gamma.
    \end{cases}
\end{equation}
where $\gamma \in (0, +\infty)$ is a hyperparameter representing the threshold at which gradients transition from being sample-sensitive to constant, it also serves as a relaxation factor, as shown in Figure \ref{F-gamma}.

Finally, the general form of the overall optimization objective is:
\begin{equation}\label{equa-L-all}
    \mathcal{L} = \mathcal{L}_{CE} +  \mathbb{E}_{v\sim \mathcal{V}} \big(\mathcal{L}_{\mathrm{NodeReg}}^v \big)
\end{equation}


\subsection{Theoretical Analysis} \label{subSec-Theory-analysis}
As an optimization term, \( \mathcal{L}_{\mathrm{NodeReg}} \) has the following properties:



\begin{property} \label{property-lipschitz}
   \textbf{\( \mathcal{L}_{\mathrm{NodeReg}} \) satisfies Lipschitz continuous}:\\ $ |\mathcal{L}_{\mathrm{NodeReg}}(\delta_1) - \mathcal{L}_{\mathrm{NodeReg}}(\delta_2)| \leq  |\delta_1 - \delta_2|, \forall \delta_1, \delta_2 \in (-\infty, 1) $
\end{property}

\begin{property} \label{property-lipschitz gradient}
   \textbf{\( \mathcal{L}_{\mathrm{NodeReg}} \) satisfies Lipschitz continuous gradient}:\\ $ |\mathcal{L}_{\mathrm{NodeReg}}^{\prime}(\delta_1) - \mathcal{L}_{\mathrm{NodeReg}}^{\prime}(\delta_2)| \leq \frac{1}{\gamma} |\delta_1 - \delta_2|, \forall \delta_1, \delta_2 \in (-\infty, 1) $
\end{property}

The Lipschitz continuity of \(\mathcal{L}_{\mathrm{NodeReg}}\) improves convergence in gradient-based methods, ensuring more reliable optimization. Besides, \(\mathcal{L}_{\mathrm{NodeReg}}\) improves robustness by maintaining controlled behavior under small input perturbations, making the model more resilient to noise. The proof of Property \ref{property-lipschitz} and Property \ref{property-lipschitz gradient} are provided in the appendix \ref{proof-lp1} and \ref{proof-lp2}, respectively.



\begin{table*}[t]
\centering
\caption{ Performance of different compared baselines with GCN as the base model.  ROS denotes a random over-sample of the nodes and their edges in minority classes to re-balance the classes.}\label{tabel-imbalance-main}
\resizebox{\linewidth}{!}{
\begin{tabular}{c|ccc|ccc|ccc|ccc|ccc}
\hline
\hline
\multirow{2}*{Method} &\multicolumn{3}{c|}{Cora} &\multicolumn{3}{c|}{Citeseer} &\multicolumn{3}{c|}{Pubmed}&\multicolumn{3}{c|}{Ogbn-arxiv}&\multicolumn{3}{c}{Amazon-computers}\\
~ &F1 score &ACC &AUC-ROC	&F1 score &ACC &AUC-ROC &F1 score &ACC &AUC-ROC &F1 score &ACC &AUC-ROC &F1 score &ACC &AUC-ROC	\\
\hline
\hline
GCN\citep{kipf2016semi} &52.9±2.1 &62.7±1.4 &87.5±0.9 &21.7±0.4 &30.5±0.5 &74.7±1.2 &51.8±2.5 &56.6±0.9 &85.2±5.9 &30.1±1.4 &37.4±0.8 &80.0±0.6 &73.9±1.1 &74.7±1.3 &96.0±0.1\\
ROS &53.0±3.8 &63.4±2.3 &85.7±1.3 &22.6±0.4 &31.0±0.3 &75.1±0.6 &61.5±0.7 &63.5±0.5 &87.3±0.9 &35.8±1.5 &40.4±0.9 &82.3±1.7 &74.5±1.2 &75.2±1.2 &96.0±0.2\\
SMOTE\citep{chawla2002smote} &53.3±3.2 &63.5±2.1 &86.1±2.0 &22.4±0.4 &30.9±0.2 &75.0±0.8 &60.6±0.9 &62.7±0.6 &87.1±0.5 &34.1±1.6 &38.7±1.6 &82.0±1.7 &73.5±0.9 &76.1±1.2 &95.7±0.8\\
Reweight\citep{ren2018learning} &55.9±1.4 &65.3±1.0 &88.1±0.5 &22.1±0.4 &30.9±0.4 &75.3±1.2 &59.4±1.2 &61.9±1.0 &88.4±0.6 &33.7±1.2 &39.4±0.6 &80.7±0.7 &76.0±0.1 &76.7±0.3 &95.9±0.3\\
DR-GCN\citep{shi2020multi} &53.3±1.6 &63.4±1.5 &83.5±1.1 &22.5±2.3 &31.2±1.8 &75.0±4.1 &57.5±7.2 &60.8±5.0 &88.3±1.8 &30.2±1.4 &38.0±1.2 &82.7±1.3 &76.1±1.3 &76.3±1.6 &95.8±0.6\\
GraphSMOTE\citep{zhao2021graphsmote} &58.9±2.0 &67.8±1.5 &87.4±0.4 &22.2±0.3 &30.7±0.4 &75.3±0.1 &68.2±0.9 &68.8±0.6 &85.4±0.3 &39.7±1.6 &41.9±1.1 &79.3±1.2 &76.7±0.9 &76.6±1.2 &\cellcolor{gray!30}96.9±0.4\\
ReNode\citep{chen2021topology} &60.3±2.3 &67.2±2.5 &88.7±1.4 &35.2±4.5 &38.4±3.7 &77.1±1.6 &68.8±1.3 &69.6±1.2 &90.3±0.2 &36.7±1.1 &41.8±1.1 &81.8±0.8 &77.2±0.6 &77.7±0.5 &96.6±0.2\\
GraphENS\citep{park2021graphens} &67.9±3.1 &72.8±1.9 &89.5±0.9 &47.9±2.8 &52.4±2.7 &80.2±0.9 &67.8±0.8 &67.1±1.2 &85.1±1.4 &38.7±1.6 &40.4±1.1 &78.1±0.6 &78.3±0.6 &79.5±0.6 &96.8±0.1\\
ReVar\citep{yan2024rethinking} &68.2±0.9 &73.3±0.7 &89.1±0.5 &48.1±0.8 &52.9±0.9 &81.0±0.5 &69.8±0.6 &70.1±0.5 &87.1±0.4 &39.9±0.5 &41.9±0.7 &80.1±0.4 &78.1±0.2 &78.9±0.2 &96.7±0.2\\
BIM\citep{zhang2024bim} &\cellcolor{gray!30}68.9±2.2 &\cellcolor{gray!30}74.0±1.9 &\cellcolor{gray!30}89.6±1.1 &\cellcolor{gray!30}48.7±7.1 &\cellcolor{gray!30}53.2±6.2 &\cellcolor{gray!30}82.0±2.7 &\cellcolor{gray!30}78.6±0.6 &\cellcolor{gray!30}79.2±0.6 &\cellcolor{gray!30}91.2±0.4 &\cellcolor{gray!30}45.6±1.9 &\cellcolor{gray!30}47.1±1.7 &\cellcolor{gray!30}84.2±1.4 &\cellcolor{gray!30}78.9±1.3 &\cellcolor{gray!30}79.7±0.1 &96.7±0.1\\
NodeReg (ours) &\cellcolor{gray!60}\textbf{71.3±0.2} &\cellcolor{gray!60}\textbf{76.6±0.7} &\cellcolor{gray!60}\textbf{91.2±0.5} &\cellcolor{gray!60}\textbf{61.3±0.6} &\cellcolor{gray!60}\textbf{64.3±0.3} &\cellcolor{gray!60}\textbf{86.5±0.4} &\cellcolor{gray!60}\textbf{79.7±0.7} &\cellcolor{gray!60}\textbf{80.1±0.5} &\cellcolor{gray!60}\textbf{91.8±0.5} &\cellcolor{gray!60}\textbf{54.4±0.3}&\cellcolor{gray!60}\textbf{55.2±0.4} &\cellcolor{gray!60}\textbf{88.0±0.3} &\cellcolor{gray!60}\textbf{80.8±0.6} &\cellcolor{gray!60}\textbf{80.6±0.5}&\cellcolor{gray!60}\textbf{97.0±0.2}\\

\hline
\hline
\end{tabular}
}
\end{table*}

\begin{table*}[t]
\centering
\caption{ Experiment results of different baselines on Cora with GCN as the base model under various imbalance ratios.}\label{tabel-imbalance-ratio}
\resizebox{\linewidth}{!}{
\begin{tabular}{c|ccc|ccc|ccc|ccc}
\hline
\hline
\multirow{3}*{Method} &\multicolumn{12}{c}{Imbalance ratios} \\
\cline{2-13}	
~ &\multicolumn{3}{c|}{0.1} &\multicolumn{3}{c|}{0.3} &\multicolumn{3}{c|}{0.5}&\multicolumn{3}{c}{0.7}\\
~ &F1 score &ACC &AUC-ROC	&F1 score &ACC &AUC-ROC &F1 score &ACC &AUC-ROC &F1 score &ACC &AUC-ROC	\\
\hline
\hline
GCN\citep{kipf2016semi} &52.9±2.1 &62.7±1.4 &87.5±0.9 &69.8±0.7 &73.3±0.6 &95.0±0.2 &76.3±0.8 &77.6±0.7 &96.1±0.1 &80.1±0.6 &81.1±0.5 &96.4±0.1\\
ROS &53.0±3.8 &63.4±2.3 &85.7±1.3 &73.2±1.1 &75.5±0.7 &94.9±0.2 &77.3±1.0 &78.4±0.8 &95.9±0.1 &80.5±0.6 &81.4±0.5 &96.5±0.1\\
SMOTE\citep{chawla2002smote} &53.3±3.2 &63.5±2.1 &86.1±2.0 &70.7±0.9 &73.7±0.8 &94.9±0.2 &77.4±0.9 &78.6±0.8 &95.9±0.2 &80.3±0.5 &81.3±0.5 &96.5±0.1\\
Reweight\citep{ren2018learning} &55.9±1.4 &65.3±1.0 &88.1±0.5 &75.4±0.9 &77.0±0.9 &95.4±0.2 &78.2±0.7 &79.4±0.6 &96.2±0.1 &80.8±0.4 &81.5±0.6 &96.7±0.1\\
DR-GCN\citep{shi2020multi} &53.3±1.6 &63.4±1.5 &83.5±1.1 &72.9±1.9 &75.4±1.8 &95.4±0.4 &78.1±1.3 &79.3±1.5 &95.9±0.1 &80.9±0.6 &81.5±0.6 &96.2±0.2\\
GraphSMOTE\citep{zhao2021graphsmote} &58.9±2.0 &67.8±1.5 &87.4±0.4 &72.8±1.4 &73.9±1.0 &91.4±0.8 &79.1±1.4 &80.0±1.3 &96.3±0.3 &80.4±0.6 &81.2±0.5 &96.3±0.2\\
ReNode\citep{chen2021topology} &60.3±2.3 &67.2±2.5 &88.7±1.4 &76.3±1.5 &78.0±1.3 &95.4±0.5 &79.1±1.1 &80.0±1.2 &96.4±0.4 &81.2±0.8 &81.8±0.8 &\cellcolor{gray!30}96.8±0.1\\
GraphENS\citep{park2021graphens} &67.9±3.1 &72.8±1.9 &89.5±0.9 &75.7±0.7 &76.8±0.4 &94.9±0.3 &79.5±0.2 &81.9±0.3 &95.5±0.4 &81.5±0.2 &82.5±0.7 &96.5±0.2\\
ReVar\citep{yan2024rethinking} &68.2±0.9 &73.3±0.7 &89.1±0.5 &76.9±0.3 &78.1±0.2 &95.2±0.2 &79.9±0.1 &81.7±0.1 &96.1±0.1 &81.8±0.2 &82.6±0.1 &96.6±0.1\\
BIM\citep{zhang2024bim} &\cellcolor{gray!30}68.9±2.2 &\cellcolor{gray!30}74.0±1.9 &\cellcolor{gray!30}89.6±1.1 &\cellcolor{gray!30}79.2±1.4 &\cellcolor{gray!30}80.2±1.5 &\cellcolor{gray!30}95.6±0.3 &\cellcolor{gray!30}81.1±0.9 &\cellcolor{gray!30}82.0±0.8 &\cellcolor{gray!30}96.4±0.3 &\cellcolor{gray!30}82.1±0.7 &\cellcolor{gray!30}82.8±0.7 &\cellcolor{gray!30}96.8±0.1\\
NodeReg (ours) &\cellcolor{gray!60}\textbf{71.3±0.2} &\cellcolor{gray!60}\textbf{76.6±0.7} &\cellcolor{gray!60}\textbf{91.2±0.5} &\cellcolor{gray!60}\textbf{80.3±0.3} &\cellcolor{gray!60}\textbf{81.1±0.6} &\cellcolor{gray!60}\textbf{96.2±0.4} &\cellcolor{gray!60}\textbf{81.5±0.8} &\cellcolor{gray!60}\textbf{82.4±0.6} &\cellcolor{gray!60}\textbf{96.6±0.3} &\cellcolor{gray!60}\textbf{82.4±0.2}&\cellcolor{gray!60}\textbf{82.9±0.3} &\cellcolor{gray!60}\textbf{97.0±0.1} \\
\hline
\hline
\end{tabular}
}
\end{table*}

\begin{table}[!t]
\centering
\caption{Datasets used in the node imbalance scenario.}\label{tabel-datasets-imbalance}
\resizebox{\linewidth}{!}{
\begin{tabular}{c|c|c|c|c|c|c}
\hline
\hline
Dataset & \# Nodes & \# Edges & \# Features & \# Classes & \makecell[c]{\# Minority\\ Classes} & \# Description \\
\hline
Cora &2,708 &5,429 &1,433 &7 &3 &citation network\\
Citeseer &3,327 &4,732 &3,703 &6 &3 &citation network\\
Pubmed &19,717 &44,338 &500 &3 &1 &citation network\\
Amazon-computers &13,752 &491,722 &767 &10 &5 &co-purchase network\\
Ogbn-arxiv &13,515 &23,801 &128 &10 &5 &citation network\\
\hline
\hline
\end{tabular}
}
\end{table}

\begin{table}[!t]
\centering
\caption{Datasets used in the OOD generalization scenario.}\label{tabel-datasets-OOD}
\resizebox{\linewidth}{!}{
\begin{tabular}{c|c|c|c|c|c}
\hline
\hline
Dataset & \# Nodes & \# Edges & \# Features & \# Classes & \# Shift Types \\
\hline
Cora &2,708 &5,429 &1,433 &7  &spurious features\\
Citeseer &3,327 &4,732 &3,703 &6  &spurious features\\
Pubmed &19,717 &44,338 &500 &3  &spurious features\\
Twitch &34,120 &892,346 &2,545 &2  &disconnected subgraphs\\
Arxiv &169,343 &1,166,243 &128 &40 &time attributes\\
\hline
\hline
\end{tabular}
}
\end{table}


\begin{proposition}\label{proposition-intra-class}
\textbf{Intra-class norm consistency minimizes overfitting to noise.}
\end{proposition}
 Recent work on benign overfitting\citep{bartlett2020benign,cao2022benign,huang2023graph} highlights a paradox between deep neural networks and traditional machine learning models. In traditional models, perfect training data fit often leads to poor test performance, while deep networks can maintain good test performance despite fitting the training data perfectly. This phenomenon has been studied from the perspective of decoupling signal and noise \citep{cao2022benign,huang2023graph}, with new methods proposed for deriving generalization bounds. We find that ensuring consistent norms within the same class prevents fitting noise, improving robustness and generalization. Experimental analysis is provided in section \ref{subSec-experiment-PopulationRisk}, with a theoretical proof in the appendix \ref{Appendix-Proof-intra-class-cnsis}.
\begin{proposition}\label{proposition-inter-class-balanced}
\textbf{Inter-class norm consistency maximizes projection variance. 
}
\end{proposition}
Recent work on neural collapse\citep{papyan2020prevalence,zhu2021geometric,zhou2022all,han2021neural,tirer2022extended} shows that during classification training, the final sample representations form a simplex equiangular tight frame projected onto a hyperplane \( \phi \in \mathbb {R}^{c-1} \). Consistent norms between classes result in the projection variance, or inter-class covariance on \( \phi \) being maximized (i.e., maximized Fisher discriminant ratio\citep{fisher1936use}), facilitating better separation between classes and thus improving the model's generalization ability. The proof of Proposition \ref{proposition-inter-class-balanced} is provided in the appendix \ref{Appendix-proposition-inter-class-balanced}.

\subsection{Other Methods for Norm Consistency} \label{subSec-method-other-NCsis}
In this section, we compare our method with other methods that also enforce norm consistency, including supervised contrastive learning loss $\mathcal{L}_{scl}$\citep{Khosla_NIPS20_SupCon}, center loss $\mathcal{L}_{center}$\citep{wen2016discriminative}, and $\mathcal{L}_{bound}$\citep{yangbounded}:
\begin{itemize}[nosep, topsep=0pt, leftmargin=*]
\item The supervised contrastive learning loss $\mathcal{L}_{scl}$ \citep{Khosla_NIPS20_SupCon} is defined as:
\begin{gather}\label{equa-loss-scl}
    \mathcal{L}_{scl} = \sum_{v_{i} \in |\mathcal{V}| } \frac{-1}{|\mathcal{V}_{y_i}|-1} \mathcal{L}_i \\
    \mathcal{L}_i =  \sum_{{v}_{j} \in |\mathcal{V}| } \mathbbm{1}_{i \neq j} \mathbbm{1}_{y_i = y_j} {\rm log} \frac{{\rm exp}({\rm s}_{i,j} /\tau)}{{\rm exp}({\rm s}_{i,j}/\tau) + {\rm exp}(\sum \limits_{v_{k} \in |\mathcal{V}|} \mathbbm{1}_{y_i \neq y_j} {\rm s}_{i,k}/\tau)} \\
    {\rm s}_{i,k}={\rm cos}<{z_i},{z_k}>={z_i}^\top{z_k}/ \lVert{z_i}\rVert \lVert{z_{k}}\rVert
\end{gather}
where $\mathbbm{1}$ is indicator function. 

\item The center loss $\mathcal{L}_{center}$ \citep{wen2016discriminative}is defined as:
\begin{equation}\label{equa-loss-center}
\mathcal{L}_{center} = \frac{1}{2} \sum_{i=1}^{m} \left\| \mathbf{z}_i - \mathbf{\mu}_{y_i} \right\|^2_2
\end{equation}
where \(m\) is the number of training samples, \(\mathbf{z}_i\) is the representation of the \(i\)-th sample, \(\mathbf{\mu}_{y_i}\) is the center of the class to which the \(i\)-th sample belongs.

\item 
The $\mathcal{L}_{bound}$ \citep{yangbounded} is defined as:
\begin{gather}\label{equa-loss-L_bound}
    M_{{norm}} = \frac{1}{|\mathcal{V}|} \sum_{v \in \mathcal{V}} \lVert \mathbf{z}_v \rVert_2 \\
    \mathcal{L}_{{bound}} = M_{{norm}}^{-1}\cdot \mathbb{E}_{\mathbf{z}\sim \mathcal{V}} \big( ( \lVert \mathbf{z} \rVert_2 - \frac{1}{|\mathcal{V}|} \sum_{v \in \mathcal{V}} \lVert \mathbf{z}_v \rVert_2)^2 \big)
\end{gather}
where $\lVert \mathbf{z}_v \rVert_2 = \sqrt{\sum_{j=0}^{c-1} \mathrm{z}^2_j}$ denotes the L2 norm of the logit $\mathbf{z}_v$. For ease of comparison, we rewrite $\mathcal{L}_{bound}$ in the following form:
\begin{gather}\label{equa-loss-L_bound-F}
    \mathcal{L}_{bound} = \bar{\mathcal{F}}^{-1}\cdot \mathbb{E}_{\mathbf{v}\sim \mathcal{V}} \big( \delta_v^2 \big)
\end{gather}

\end{itemize}

Compared to \(\mathcal{L}_{scl}\)  and \(\mathcal{L}_{center} \), our method has the following two advantages: \textbf{1. No Need for Label Supervision.} Our method operates without the need for labels, enabling the application of the norm consistency constraint to unlabeled node representations. This promotes global norm consistency throughout the entire graph, allowing the norms of node representations across different classes to align. Such consistency is particularly advantageous for semi-supervised node classification in cases with imbalanced nodes. In contrast, \(\mathcal{L}_{scl}\)  and \(\mathcal{L}_{center} \) cannot achieve this.
\textbf{2. Focus on Norm Optimization.} While \(\mathcal{L}_{scl}\)  and \(\mathcal{L}_{center} \) optimize the angles and norms between representations, our method focuses on norm optimization. This results in better global representation norm consistency and higher time efficiency. 

Compared to $\mathcal{L}_{bound}$, our method satisfies Lipschitz continuity, while $\mathcal{L}_{bound}$ does not, as shown below:
\begin{equation}\label{equa-loss-L_bound-F-Gradient}
    \frac{d}{d \delta}\mathcal{L}_{bound}^v = 2\bar{\mathcal{F}}^{-1}\cdot \delta_v 
\end{equation}
By computing the derivative of $\mathcal{L}_{bound}^v$, we observe that the derivative depends on $\delta_v$, so there is no constant upper bound. This leads to a less stable optimization of $\mathcal{L}_{bound}^v$ compared to our method, especially in the early stages of training, where the original NODESAFE\citep{yangbounded} paper mentions the need for dozens of training epochs before $\mathcal{L}_{bound}^v$ can be used. Consequently, this instability can cause the training process to struggle to converge to the optimal performance, particularly in scenarios with node imbalance.

\section{Experiments}

We apply our method to alleviate the problems of node imbalance and OOD generalization in semi-supervised node classification. Our research aims to address the following questions:
\begin{itemize}[nosep, topsep=0pt, leftmargin=*]
\item \textbf{RQ1}: How does NodeReg perform compared to state-of-the-art methods in scenarios with node imbalance?

\item \textbf{RQ2}: How does NodeReg perform compared to state-of-the-art methods in scenarios with node distribution shift?


\item \textbf{RQ3}: How does NodeReg perform compared to other loss functions that promote representation norm consistency?


\item \textbf{RQ4}: Can NodeReg reduce GNN's overfitting to noise?
\item \textbf{RQ5}: How different $\gamma$ affects the performance of NodeReg?
\end{itemize}

\subsection{Experiment Setup}
We follow the experimental settings of BIM \citep{zhang2024bim} for node imbalance and CaNet \citep{wu2024graph} for OOD generalization. Details and baseline methods are provided in the appendix \ref{Sec-Appendix-Exp-Setting}.

 \subsubsection{Datasets} 
\textbf{Node Imbalance Scenario.} We evaluate the performance of various algorithms using several real-world networks in line with BIM\citep{zhang2024bim}: Ogbn-Arxiv\citep{hu2020open}, Cora, Citeseer, PubMed\citep{kipf2016semi}, and Amazon-Computers\citep{shchur2018pitfalls}. 
    We use the GCN\citep{kipf2016semi} public split for Cora, Citeseer, and Pubmed, widely used citation networks for node property prediction. In the Amazon-Computers dataset, nodes represent products, while edges indicate frequent co-purchases. The task is to classify products using bag-of-words features derived from product reviews. For Ogbn-Arxiv, feature vectors for papers are created by averaging word embeddings from abstracts. Following the BIM\citep{zhang2024bim}, we focus on the ten largest classes from each dataset to maintain experimental consistency. The majority class in the training set contains 20 nodes, and minority classes contain 20 multiplied by the imbalance ratio (default 0.1). Table \ref{tabel-datasets-imbalance} provides a summary of the dataset characteristics.



\begin{table*}[!t]
\centering
\caption{Test (mean ± standard deviation) Accuracy (\%) for citation networks on out-of-distribution (OOD) and in-distribution (ID)
data. The distribution shifts are introduced by generating different node features (OOD-Feat) and graph structures (OOD-Struct).
OOM indicates an out-of-memory error on a GPU with 24GB of memory.}\label{tabel-OOD-Cora-Citeseer-Pubmed}
\resizebox{\linewidth}{!}{
\begin{tabular}{c|cc|cccccccccc}
\hline
\hline
Backbone & \multicolumn{2}{c|}{Dataset}  & ERM & IRM\citep{arjovsky2019invariant} & Deep Coral\citep{sun2016deep} & DANN\citep{ganin2016domain} & GroupDRO\citep{sagawa2019distributionally} & Mixup\citep{zhang2017mixup} & SRGNN\citep{zhu2021shift} & EERM\citep{wu2022handling} & CaNet\citep{wu2024graph} &  NodeReg (ours)\\
\hline
\multirow{6}{*}{\rotatebox{90}{GCN\citep{kipf2016semi}}}&\multirow{2}{*}{Cora} & OOD & 74.30 ± 2.66 & 74.19 ± 2.60 & 74.26 ± 2.28 & 73.09 ± 3.24 & 74.25 ± 2.61 & 92.77 ± 1.27 & 81.91 ± 2.64 & 83.00 ± 0.77 & \cellcolor{gray!30}96.12 ± 1.04 & \cellcolor{gray!60}\textbf{98.64 ± 0.21} \\
~&~ & ID & 94.83 ± 0.25 & 94.88 ± 0.18 & 94.89 ± 0.18 & 95.03 ± 0.16 & 94.87 ± 0.25 & 94.84 ± 0.30 & 95.09 ± 0.32 & 89.17 ± 0.23 & 97.87 ± 0.23 & \textbf{97.92 ± 0.19} \\
\cline{2-13}
~&\multirow{2}{*}{Citeseer} & OOD & 74.93 ± 2.39 & 75.34 ± 1.61 & 74.97 ± 2.53 & 74.74 ± 2.78 & 75.02 ± 2.05 & 77.28 ± 5.28 & 76.10 ± 4.04 & 74.76 ± 1.15 & \cellcolor{gray!30}94.57 ± 1.92 & \cellcolor{gray!60}\textbf{97.47 ± 0.13} \\
~&~&  ID & 85.76 ± 0.26 & 85.34 ± 0.46 & 85.64 ± 0.28 & 85.75 ± 0.49 & 85.33 ± 0.36 & 85.00 ± 0.50 & 85.84 ± 0.37 & 83.81 ± 0.17 & 95.18 ± 0.17 & \textbf{95.24 ± 0.07} \\
\cline{2-13}
~&\multirow{2}{*}{Pubmed} & OOD & 81.36 ± 1.78 & 81.14 ± 1.72 & 81.56 ± 2.35 & 80.77 ± 1.43 & 81.07 ± 1.89 & 79.76 ± 4.44 & 84.75 ± 2.38 & OOM & \cellcolor{gray!30}88.82 ± 2.30 & \cellcolor{gray!60}\textbf{90.55 ± 0.66} \\
~&~&  ID & 92.76 ± 0.10 & 92.80 ± 0.12 & 92.78 ± 0.11 & 93.20 ± 0.42 & 92.76 ± 0.08 & 92.68 ± 0.13 & 93.52 ± 0.31 & OOM & 97.37 ± 0.11 & \textbf{97.53 ± 0.06} \\
\hline
\multirow{6}{*}{\rotatebox{90}{GAT\citep{2017Graph}}}&\multirow{2}{*}{Cora} & OOD & 91.10 ± 2.26 & 91.63 ± 1.27 & 91.82 ± 1.30 & 92.40 ± 2.05 & 90.54 ± 0.94 & 92.94 ± 1.21 & 91.77 ± 2.43 & 91.80 ± 0.73 & \cellcolor{gray!30}97.30 ± 0.25 & \cellcolor{gray!60}\textbf{98.54 ± 0.08} \\
 ~&~& ID & 95.57 ± 0.40 & 95.72 ± 0.31 & 95.74 ± 0.39 & 95.66 ± 0.28 & 95.38 ± 0.23 & 94.66 ± 0.10 & 95.36 ± 0.24 & 91.37 ± 0.30 & 95.94 ± 0.29 & \textbf{97.35 ± 0.05} \\
\cline{2-13}
~&\multirow{2}{*}{Citeseer} & OOD & 82.60 ± 0.51 & 82.73 ± 0.37 & 82.44 ± 0.58 & 82.49 ± 0.67 & 82.64 ± 0.61 & 82.77 ± 0.30 & 82.72 ± 0.35 & 74.07 ± 0.75 & \cellcolor{gray!30}95.33 ± 0.33 & \cellcolor{gray!60}\textbf{96.56 ± 0.09} \\
~&~ & ID & 89.02 ± 0.32 & 89.11 ± 0.36 & 89.05 ± 0.37 & 89.02 ± 0.31 & 89.13 ± 0.27 & 89.05 ± 0.05 & 89.10 ± 0.15 & 83.53 ± 0.56 & 89.57 ± 0.65 & \textbf{91.81 ± 0.45} \\
\cline{2-13}
~&\multirow{2}{*}{Pubmed} & OOD & 84.80 ± 1.47 & 84.95 ± 1.06 & 85.07 ± 0.95 & 83.94 ± 0.84 & 85.17 ± 0.86 & 81.58 ± 0.65 & 83.40 ± 0.67 & OOM & \cellcolor{gray!30}89.89 ± 1.92 & \cellcolor{gray!60}\textbf{91.96 ± 0.97} \\
 ~&~& ID & 93.98 ± 0.24 & 93.89 ± 0.26 & 94.05 ± 0.23 & 93.46 ± 0.31 & 94.00 ± 0.18 & 92.79 ± 0.18 & 93.21 ± 0.29 & OOM & 95.04 ± 0.16 & \textbf{95.89 ± 0.24} \\
\hline
\hline
\end{tabular}
}
\end{table*}

\begin{table*}[!t]
\centering
\caption{Test (mean ± standard deviation) Accuracy (\%) for Arxiv and ROC-AUC (\%) for Twitch on different subsets of out-of-distribution data. We use publication years and subgraphs for data splits on Arxiv and Twitch, respectively.}\label{tabel-OOD-Arxiv-Twitch-horizontal}
\resizebox{\linewidth}{!}{
\begin{tabular}{c|cc|cccccccccc}
\hline
\hline
 Backbone & \multicolumn{2}{c|}{Dataset}  & ERM & IRM\citep{arjovsky2019invariant} & Deep Coral\citep{sun2016deep} & DANN\citep{ganin2016domain} & GroupDRO\citep{sagawa2019distributionally} & Mixup\citep{zhang2017mixup} & SRGNN\citep{zhu2021shift} & EERM\citep{wu2022handling} & CaNet\citep{wu2024graph} &  NodeReg (ours) \\
\hline
\multirow{8}{*}{\rotatebox{90}{GCN\citep{kipf2016semi}}}&\multirow{4}{*}{Arxiv} 
& OOD1 & 56.33 ± 0.17 & 55.92 ± 0.24 & 56.42 ± 0.26 & 56.35 ± 0.11 & 56.52 ± 0.27 & 56.67 ± 0.46 & 56.79 ± 1.35 & OOM & \cellcolor{gray!30}59.01 ± 0.30 & \cellcolor{gray!60}\textbf{59.85 ± 0.52} \\
~&~& OOD2 & 53.53 ± 0.44 & 53.25 ± 0.49 & 53.53 ± 0.54 & 53.81 ± 0.33 & 53.40 ± 0.29 & 54.02 ± 0.51 & 54.33 ± 1.78 & OOM & \cellcolor{gray!60}\textbf{56.88 ± 0.70} & \cellcolor{gray!30}56.47 ± 1.47 \\
~&~& OOD3 & 45.83 ± 0.47 & 45.66 ± 0.83 & 45.92 ± 0.52 & 45.89 ± 0.37 & 45.76 ± 0.59 & 46.09 ± 0.58 & 46.24 ± 1.90 & OOM & \cellcolor{gray!30}56.27 ± 1.21 & \cellcolor{gray!60}\textbf{56.53 ± 1.55} \\
~&~& ID & 59.94 ± 0.45 & 60.28 ± 0.23 & 60.16 ± 0.12 & 60.22 ± 0.29 & 60.35 ± 0.27 & 60.09 ± 0.15 & 60.02 ± 0.52 & OOM & 61.42 ± 0.10 & \textbf{61.45 ± 0.15} \\
\cline{2-12}
~&\multirow{4}{*}{Twitch} 
& OOD1 & 66.07 ± 0.14 & 66.95 ± 0.27 & 66.15 ± 0.14 & 66.15 ± 0.13 & 66.82 ± 0.26 & 65.76 ± 0.30 & 65.83 ± 0.45 & 67.50 ± 0.74 & \cellcolor{gray!30}67.47 ± 0.32 & \cellcolor{gray!60}\textbf{67.90 ± 0.15} \\
~&~& OOD2 & 52.62 ± 0.01 & 52.53 ± 0.02 & 52.67 ± 0.02 & 52.66 ± 0.02 & 52.69 ± 0.02 & 52.78 ± 0.04 & 52.47 ± 0.06 & 51.88 ± 0.07 & \cellcolor{gray!30}53.59 ± 0.19 & \cellcolor{gray!60}\textbf{53.81 ± 0.24} \\
~&~& OOD3 & 63.15 ± 0.08 & 62.91 ± 0.08 & 63.18 ± 0.03 & 63.20 ± 0.06 & 62.95 ± 0.11 & 63.15 ± 0.08 & 62.74 ± 0.23 & 62.56 ± 0.02 & \cellcolor{gray!30}64.24 ± 0.18 & \cellcolor{gray!60}\textbf{64.52 ± 0.21} \\
~&~& ID & 75.40 ± 0.01 & 74.88 ± 0.02 & 75.40 ± 0.01 & 75.40 ± 0.02 & 75.03 ± 0.01 & 75.47 ± 0.06 & 75.75 ± 0.09 & 74.85 ± 0.05 & \textbf{75.10 ± 0.08} & 75.08 ± 0.11 \\
\hline
\multirow{8}{*}{\rotatebox{90}{GAT\citep{2017Graph}}}&\multirow{4}{*}{Arxiv}
& OOD1 & 57.15 ± 0.25 & 56.55 ± 0.18 & 57.40 ± 0.51 & 57.23 ± 0.18 & 56.69 ± 0.27 & 57.17 ± 0.33 & 56.69 ± 0.38 & OOM & \cellcolor{gray!60}\textbf{60.44 ± 0.27} & \cellcolor{gray!30}59.96 ± 0.24 \\
~&~& OOD2 & 55.07 ± 0.58 & 54.53 ± 0.32 & 55.14 ± 0.71 & 55.13 ± 0.46 & 54.51 ± 0.49 & 55.33 ± 0.37 & 55.01 ± 0.55 & OOM & \cellcolor{gray!30}58.54 ± 0.72 & \cellcolor{gray!60}\textbf{59.24 ± 0.33} \\
~&~& OOD3 & 46.22 ± 0.82 & 46.01 ± 0.33 & 46.71 ± 0.61 & 46.61 ± 0.57 & 46.00 ± 0.59 & 47.17 ± 0.84 & 46.88 ± 0.58 & OOM & \cellcolor{gray!30}59.61 ± 0.28 & \cellcolor{gray!60}\textbf{60.44 ± 0.20} \\
~&~& ID & 59.72 ± 0.35 & 59.94 ± 0.18 & 60.59 ± 0.30 & 59.72 ± 0.14 & 60.03 ± 0.32 & 59.84 ± 0.50 & 59.39 ± 0.17 & OOM & 62.91 ± 0.35 & \textbf{62.97 ± 0.35} \\
\cline{2-12}
~&\multirow{4}{*}{Twitch}
& OOD1 & 65.67 ± 0.02 & 67.27 ± 0.19 & 67.12 ± 0.03 & 66.59 ± 0.38 & 67.41 ± 0.04 & 65.58 ± 0.13 & 66.17 ± 0.03 & 66.80 ± 0.46 & \cellcolor{gray!30}68.08 ± 0.19 & \cellcolor{gray!60}\textbf{68.26 ± 0.20} \\
~&~& OOD2 & 52.00 ± 0.10 & 52.85 ± 0.15 & 52.61 ± 0.01 & 52.88 ± 0.12 & 52.99 ± 0.08 & 52.04 ± 0.04 & 52.84 ± 0.04 & 52.39 ± 0.20 & \cellcolor{gray!30}53.49 ± 0.14 & \cellcolor{gray!60}\textbf{53.57 ± 0.17} \\
~&~& OOD3 & 61.85 ± 0.05 & 62.40 ± 0.24 & 63.41 ± 0.01 & 62.47 ± 0.32 & 62.29 ± 0.03 & 61.75 ± 0.13 & 62.07 ± 0.04 & 62.07 ± 0.68 & \cellcolor{gray!30}63.76 ± 0.17 & \cellcolor{gray!60}\textbf{63.94 ± 0.15} \\
~&~& ID & 75.75 ± 0.15 & 75.30 ± 0.09 & 75.20 ± 0.01 & 75.82 ± 0.27 & 75.74 ± 0.02 & 75.72 ± 0.07 & 75.45 ± 0.03 & 75.19 ± 0.50 & \textbf{76.14 ± 0.07} & 76.11 ± 0.08 \\
\hline
\hline
\end{tabular}
}
\end{table*}

\textbf{OOD Generalization Scenario.} 
We use five datasets: Cora, Citeseer, Pubmed, Twitch, and Arxiv in line with CaNet\citep{wu2024graph}. Table \ref{tabel-datasets-OOD} presents the experimental dataset statistics.
 Following the CaNet\citep{wu2024graph}, we synthetically generate spurious node features to introduce distribution shifts. 
 Arxiv is a temporal citation network \citep{hu2020open}, where papers from 2005-2014 are used as ID data, and those after 2014 as OOD data.
Twitch is a multi-graph dataset\citep{rozemberczki2021twitch} consisting of social networks from different regions. Data splits are based on subgraphs, as they differ in size, density, and node degrees. We use nodes from the DE, PT, and RU subgraphs as ID data and nodes from the ES, FR, and EN subgraphs as OOD data.

\subsubsection{Evaluation Metrics} 
\textbf{Node Imbalance Scenario.} We follow BIM\citep{zhang2024bim}, which uses classification accuracy (ACC), AUC-ROC, and F1 score for evaluation, with AUC-ROC and F1 score calculated as non-weighted averages across classes. We run the experiment
for each case using different initialization five times and reporting the means and standard deviations for the metric.

\textbf{OOD Generalization Scenario.}  We follow CaNet\citep{wu2024graph}, where ID data is split into 50\%/25\%/25\% for training/validation/testing. We assess models on ID and OOD data, focusing on OOD generalization. Accuracy is used for most datasets, ROC-AUC for Twitch. We run the experiment
for each case using different initialization five times and reporting the means and standard deviations for the metric.

\subsubsection{Implementations and Settings} All methods use the same GNN model with tuned hyper-parameters. Experiments run on Ubuntu 20.04, Cuda 12.1, Pytorch 1.12.0, and Pytorch Geometric 2.1.0.post1, using an NVIDIA 3090 with 24GB memory.

\begin{table}[!t]
\centering
\caption{Comparison of different methods that promote consistency of node representation norms in node imbalance scenarios. We combine these methods with BIM and compare the F1-score(↑) and running time (s).}\label{tabel-LOSS-Imbalance}
\resizebox{\linewidth}{!}{
\begin{tabular}{c|cc|cc}
\hline
\hline
\multirow{2}*{Method} &\multicolumn{2}{c|}{Cora} &\multicolumn{2}{c}{Citeseer} \\
~ &F1 score	&Time (s)	&F1 score	&Time (s) 			\\
\hline
\hline
BIM & 68.9 ± 2.2 &56.1 ± 1.7 &48.7 ± 7.1 &49.6 ± 1.4\\   
w/ $\mathcal{L}_{scl}$ & 58.5 ± 1.5 &90.6 ± 5.1 &39.0 ± 4.5 &88.7 ± 4.3\\   
w/ $\mathcal{L}_{center}$ & 62.4 ± 1.9 &106.4 ± 7.2 &41.5 ± 4.1 &100.7 ± 7.7\\   
w/ $\mathcal{L}_{bound}$ & 48.1 ± 2.1 &57.1 ± 2.7 &35.4 ± 5.3 &50.0 ± 1.8\\   
w/ $\mathcal{L}_{\mathrm{NodeReg}}$(ours) & 71.3 ± 0.2 &56.8 ± 1.7 &61.3 ± 0.6 &49.7 ± 2.3\\   
\hline
\hline
\end{tabular}
}
\end{table}

\begin{figure*}[!t]
	\centering
   \subfigure[Training w/o norm consistency]{
		\includegraphics[width=0.185\linewidth]{Figures/GCN_with_celoss_train_solid3.png}
 }
    \subfigure[Training w/ $\mathcal{L}_{scl}$]{
		\includegraphics[width=0.185\linewidth]{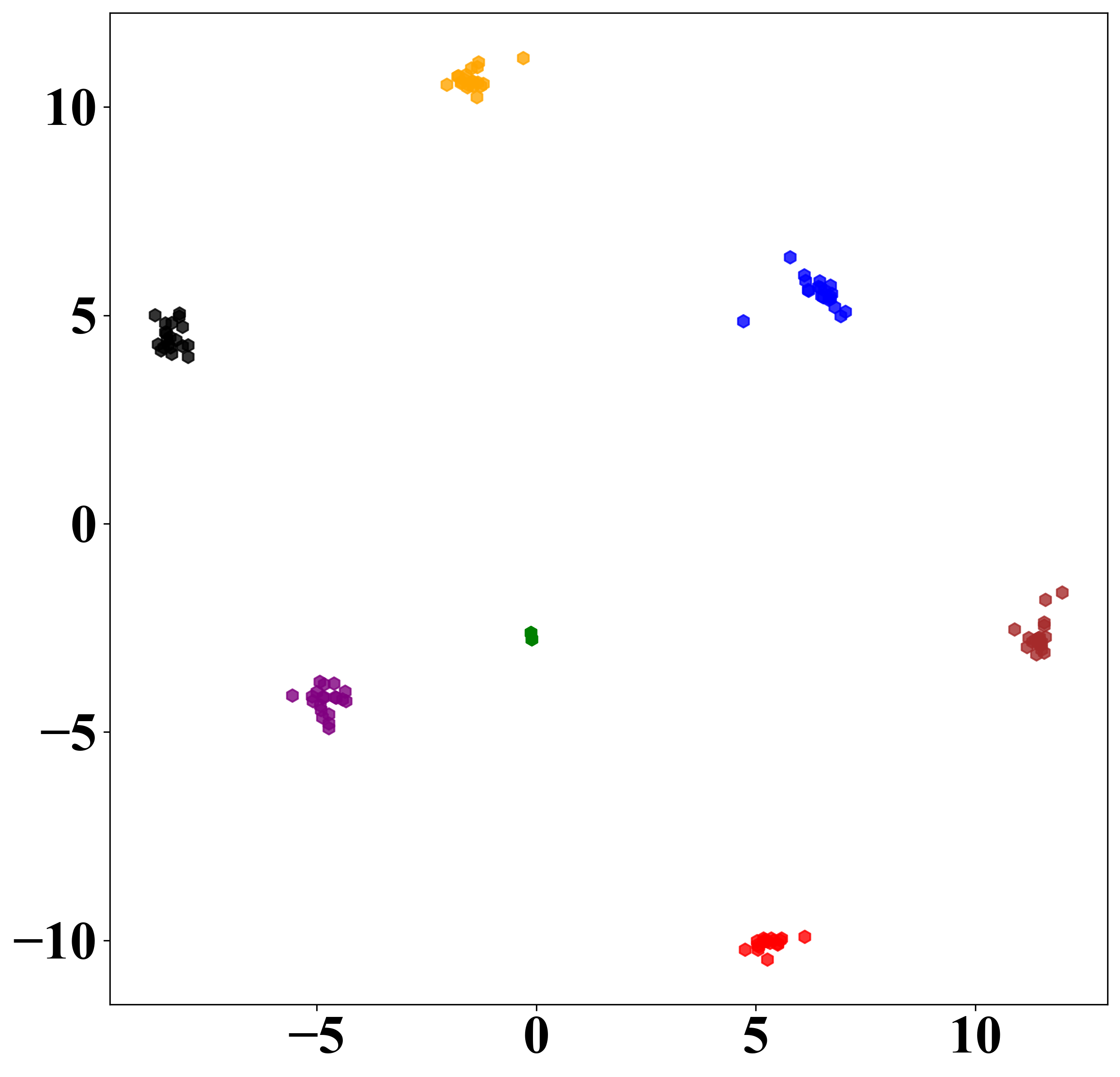}
 }
     \subfigure[Training w/ $\mathcal{L}_{center}$]{
		\includegraphics[width=0.185\linewidth]{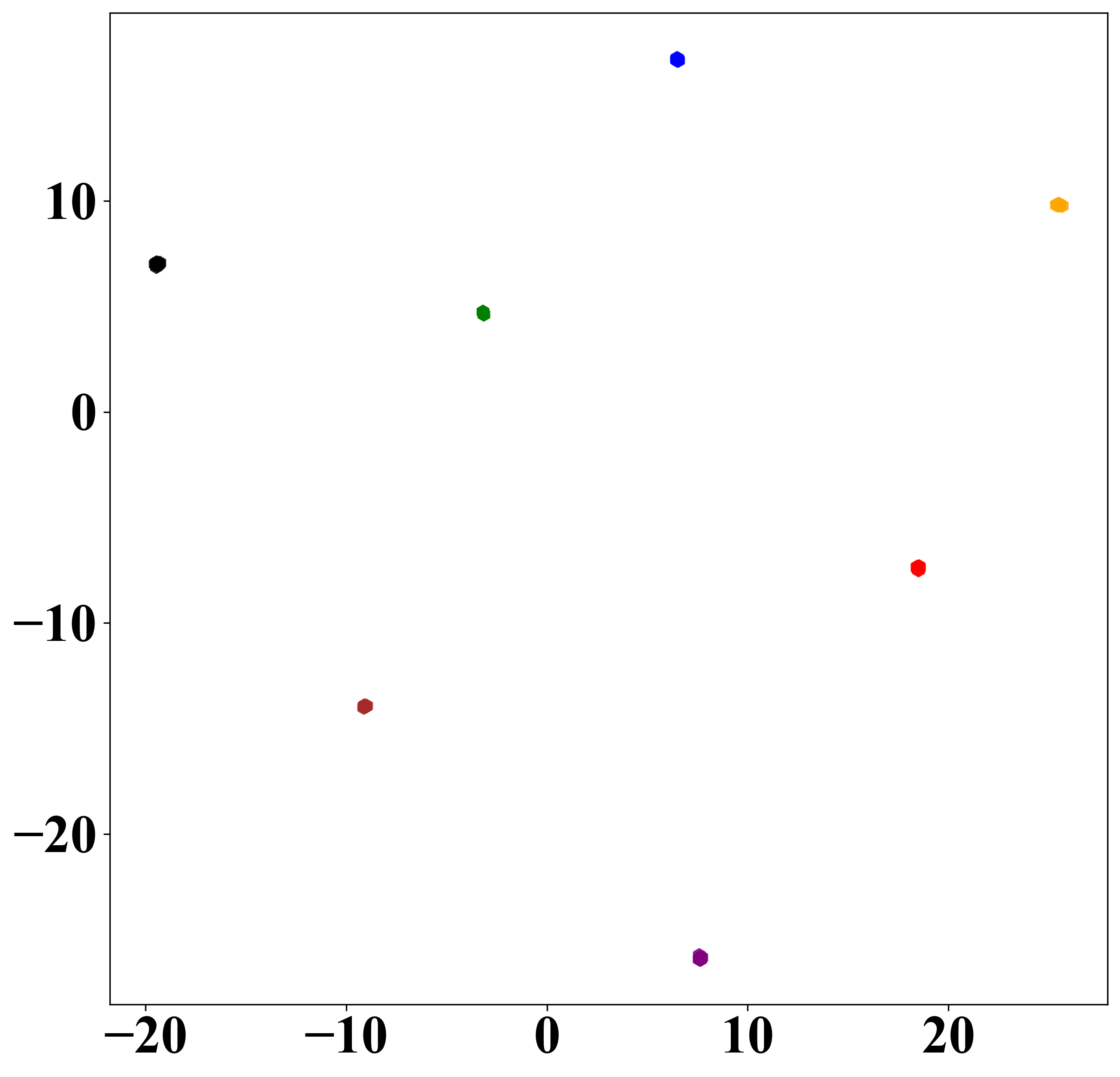}
 }
     \subfigure[Training w/ $\mathcal{L}_{bound}$]{
		\includegraphics[width=0.185\linewidth]{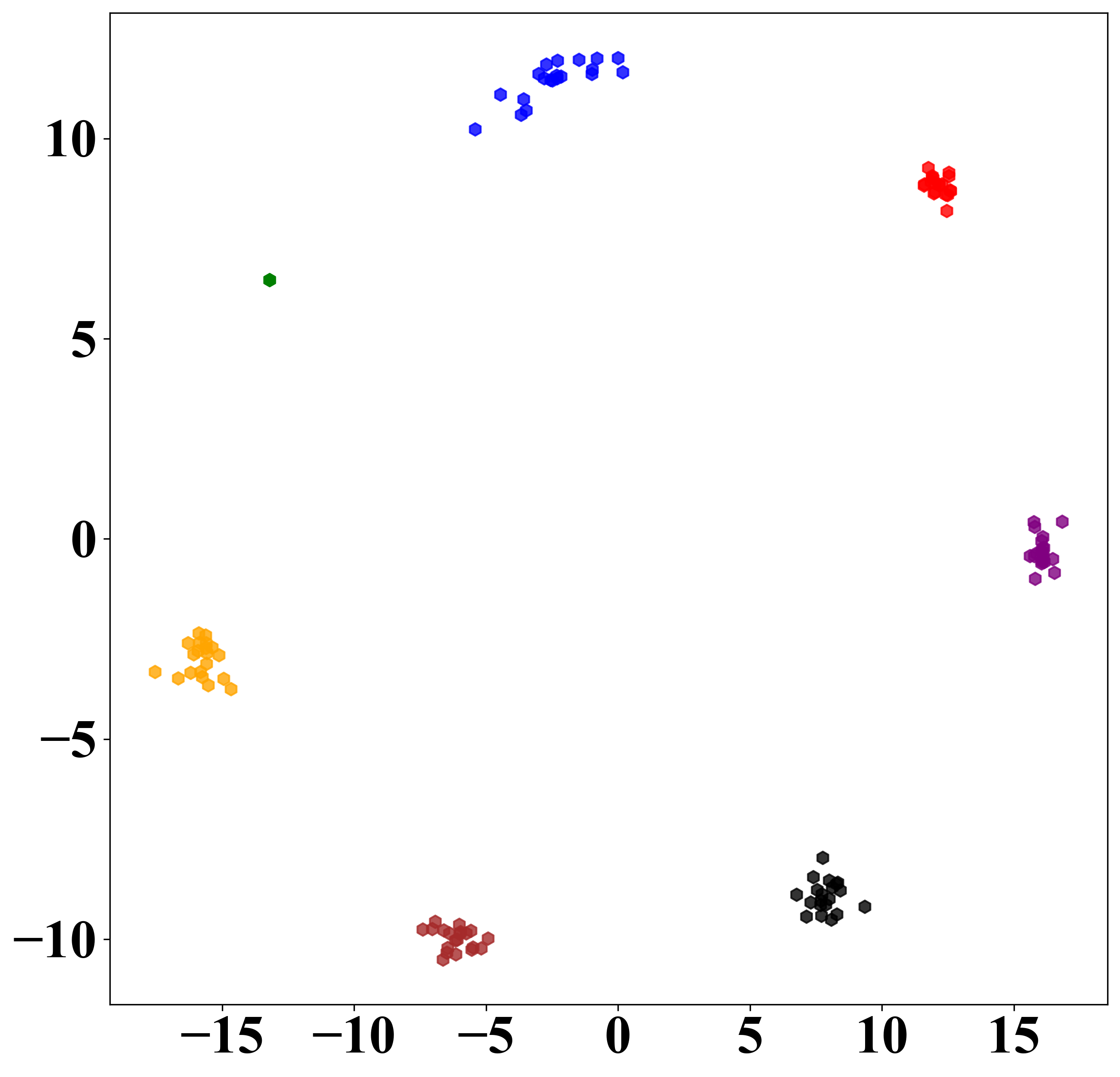}
}
    \subfigure[Training w/ $\mathcal{L}_{\mathrm{NodeReg}}$]{
		\includegraphics[width=0.185\linewidth]{Figures/GCN_with_celoss_enloss_train_solid_test.png}
 }
 
  \subfigure[Testing w/o norm consistency ]{
		\includegraphics[width=0.185\linewidth]{Figures/GCN_with_celoss_test_solid3.png}
 }
    \subfigure[Testing w/ $\mathcal{L}_{scl}$]{
		\includegraphics[width=0.185\linewidth]{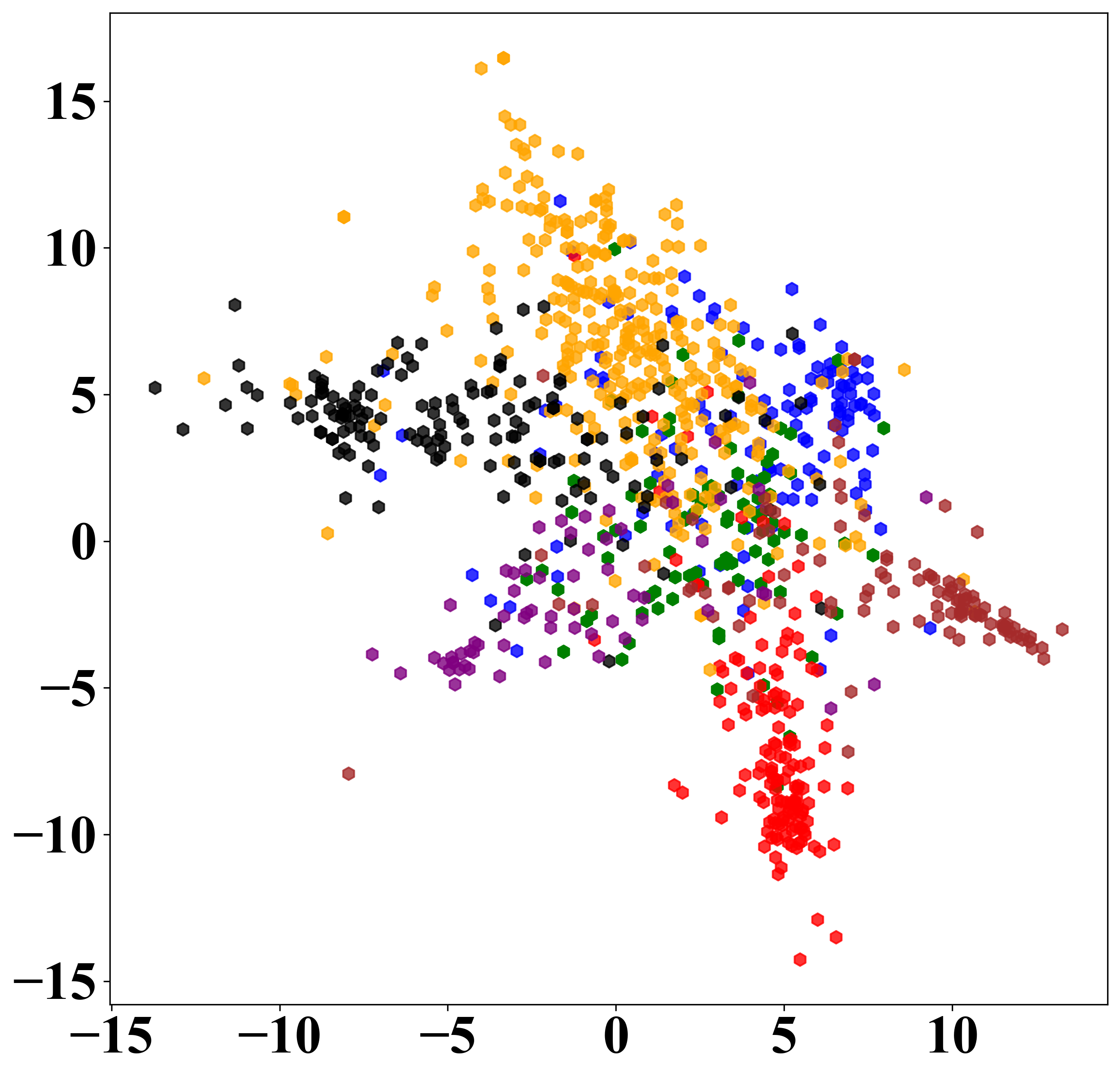}
 }    
    \subfigure[Testing w/ $\mathcal{L}_{center}$]{
		\includegraphics[width=0.185\linewidth]{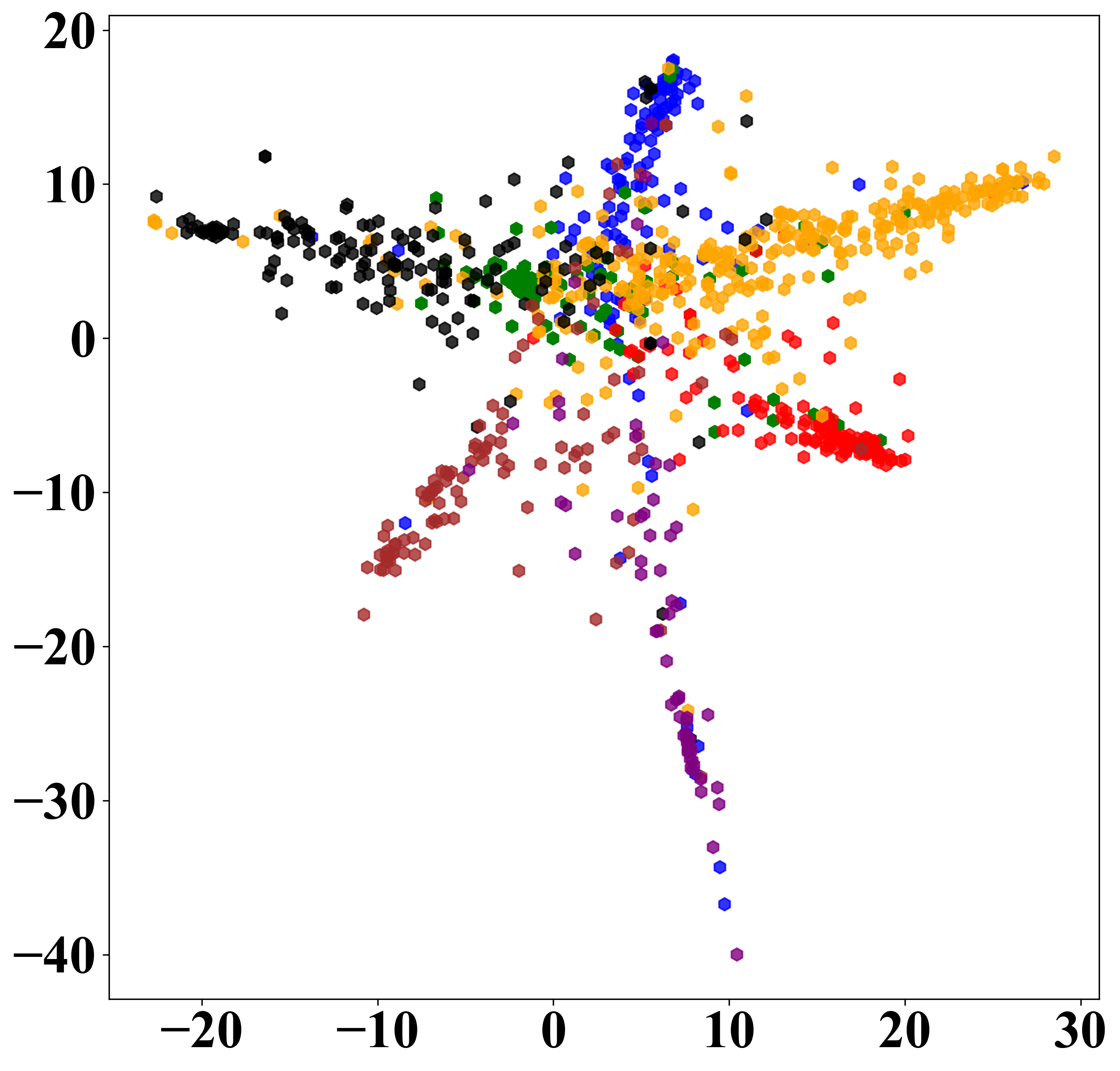}
 }    
     \subfigure[Testing w/ $\mathcal{L}_{bound}$]{
		\includegraphics[width=0.185\linewidth]{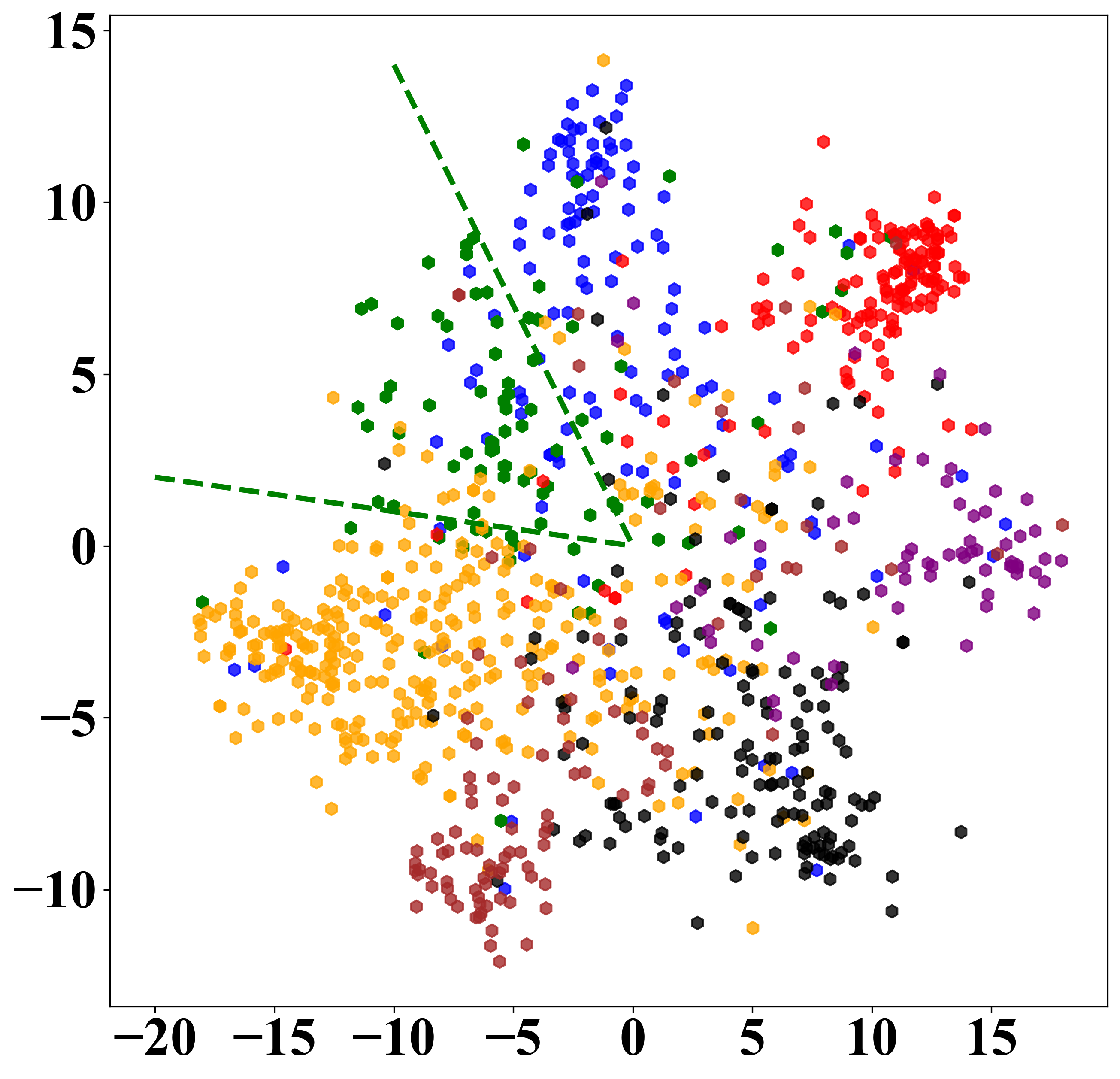}
 }    
    \subfigure[Testing w/ $\mathcal{L}_{\mathrm{NodeReg}}$]{
		\includegraphics[width=0.185\linewidth]{Figures/GCN_with_celoss_enloss_test_solid_test.png}
 }
 \Description{The visualizations of the node representations from the training set (top) and the corresponding test set (bottom) obtained by training with different norm-constrained loss functions.}
    \caption{The visualizations of the node representations from the training set (top) and the corresponding test set (bottom) obtained by training with different norm-constrained loss functions. We use GCN as the backbone.}
    \label{F-Visual}
\end{figure*}

\subsection{Comparative Results}
\subsubsection{Node Imbalance Scenario (RQ1)}
From a norm perspective, we observe that during training, minority class node representations tend to have much smaller norms compared to majority class nodes, leading to norm imbalance. This imbalance results in minority class nodes of the test set having smaller norms, making them harder to distinguish from majority class nodes.
The results in Table \ref{tabel-imbalance-main} demonstrate that when training a GCN with an extreme imbalance rate of 0.1, NodeReg's F1 scores, when combined with BIM, show significant improvements across the five public datasets. Specifically, NodeReg outperforms the most competitive baseline by a margin of 1.4\% to 25.9\%, with an average accuracy improvement of 8.8\% and an average ROC-AUC increase of 2.3\%.
To verify the effectiveness of our method under different levels of imbalance, we conducted experiments on the Cora dataset with node imbalance ratios of \{0.1, 0.3, 0.5, 0.7\}. The results in Table \ref{tabel-imbalance-ratio} show that our method consistently achieves the best performance.

\subsubsection{OOD Generalization Scenario (RQ2)}
From a norm-based perspective, we observe that node representations from different classes tend to overlap in the small-norm region, making them vulnerable to distribution shifts and leading to misclassifications. NodeReg enforces norm consistency, effectively separating overlapping representations and improving the model's robustness to distribution shifts.
In the OOD generalization scenario, NodeReg significantly outperforms the most competitive baseline by 1.4\%-3.1\% in accuracy. Table \ref{tabel-OOD-Cora-Citeseer-Pubmed} shows that our method achieves consistent and significant improvements in OOD scenarios where node features are shifted. In Table \ref{tabel-OOD-Arxiv-Twitch-horizontal}, our method achieves the best results in most experiments under the time-based OOD and subgraph OOD scenarios. However, the improvements in the time-based and subgraph OOD scenarios are not as significant as those in the node feature shift OOD scenario. We attribute this to several reasons: 
Firstly, the semi-supervised classification performance of GCN on the Arxiv dataset is suboptimal. We believe GCN tends to overfit noise even signal less on Arxiv because of the insufficient number of parameters of ordinary GCN compared to Arxiv which is large, limiting the improvement from our method, which serves as a regularizer against overfitting. Secondly, the modest improvement in the subgraph OOD scenario may be due to the fact that node representations from discrete OOD subgraphs are unseen during training, making it difficult for our method to enforce norm consistency. Consequently, the enhancement is limited in this case. In future work, we aim to refine our approach to better address these scenarios and further improve OOD generalization on GNNs.



\subsection{Comparison with Different Methods for Norm Consistency (RQ3)}\label{subSec-experiment-compare-other-NCsis}
In this section, we compare our method with several other methods that achieve varying degrees of consistency of node representation norms in the node imbalance scenario. The experiments in the OOD generalization scenario are provided in the appendix \ref{appendx-subsection-comparison-OOD-loss}.

\textbf{Comparison with $\mathcal{L}_{scl}$.} 
Figure \ref{F-Visual}(b) visualizes the node representations from the training set obtained using $\mathcal{L}_{scl}$. Compared to Figure \ref{F-Visual}(e), which depicts the node representations generated by our method, the inter-class consistency is notably weaker with $\mathcal{L}_{scl}$. In particular, the minority class nodes (green) continue to suffer from excessively small norms. As a result, Figure \ref{F-Visual}(g), which shows the test set node representations, reveals significant overlap between the minority class nodes and the majority classes. This observation is further supported by the experimental results in Table \ref{tabel-LOSS-Imbalance}, which confirm that $\mathcal{L}_{scl}$ is less effective than our method. Besides, NodeReg is also more time-efficient than $\mathcal{L}_{scl}$.

\textbf{Comparison with $\mathcal{L}_{center}$.} Figure \ref{F-Visual}(c) shows the visualization of training set node representations obtained using $\mathcal{L}_{center}$. Compared to NodeReg, $\mathcal{L}_{center}$ achieves very high intra-class consistency, but the inter-class norm consistency is not as good as ours. It also has the same problem as $\mathcal{L}_{scl}$, where significant overlap exists between the minority class node representations and majority classes in the test set. Therefore, as shown in Table \ref{tabel-LOSS-Imbalance}, its performance is not as good as our method. Besides, $\mathcal{L}_{center}$ is time-consuming because it involves operations for calculating and updating class centers, which our method does not require. As a result, our method is far more time-efficient than $\mathcal{L}_{center}$

\begin{figure}[!t]
	\centering
   \subfigure[Training w/o $\mathcal{L}_{\mathrm{NodeReg}}$]{
		\includegraphics[width=0.39\linewidth]{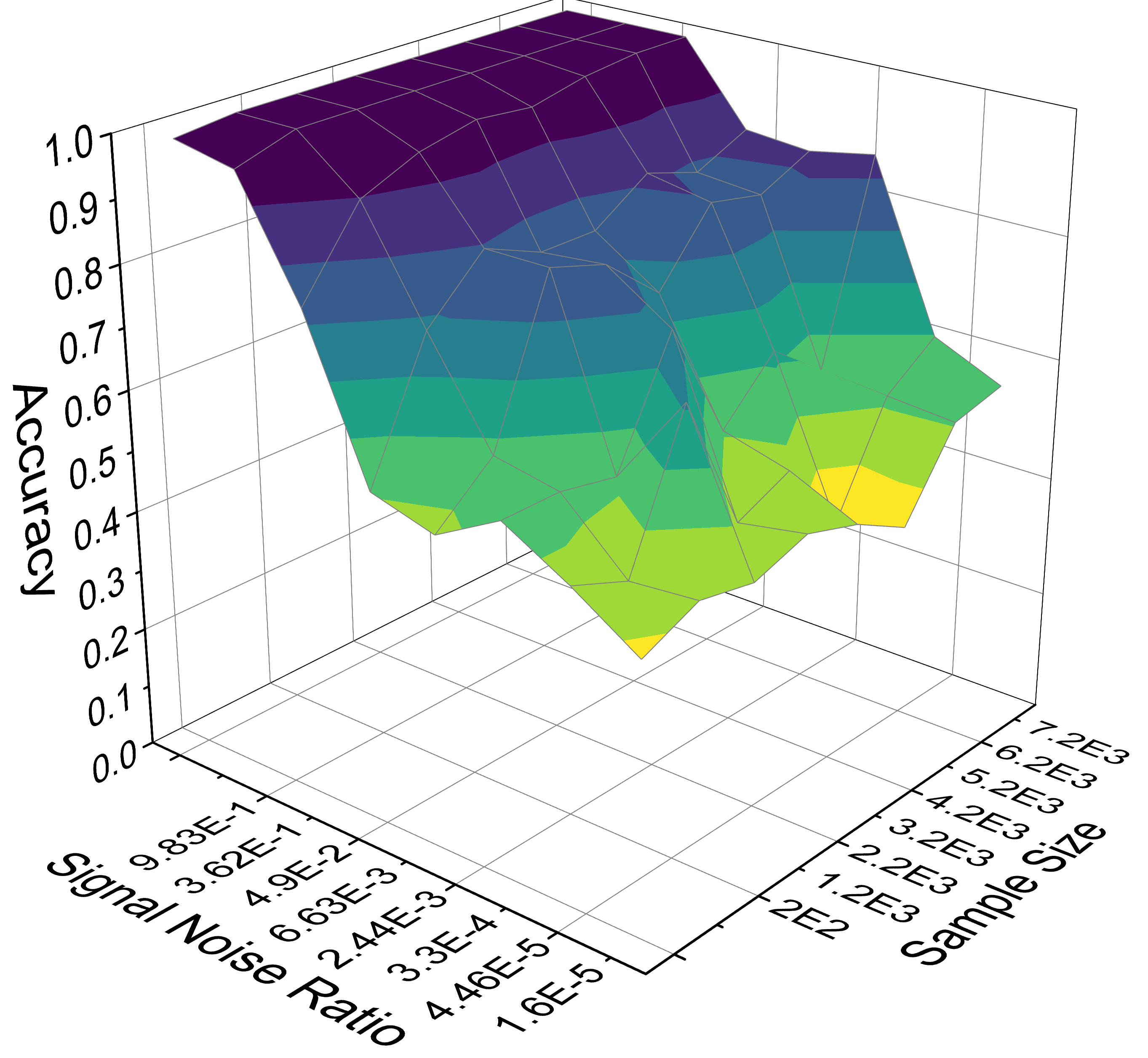}
 }
   \subfigure[Training w/ $\mathcal{L}_{\mathrm{NodeReg}}$]{
		\includegraphics[width=0.45\linewidth]{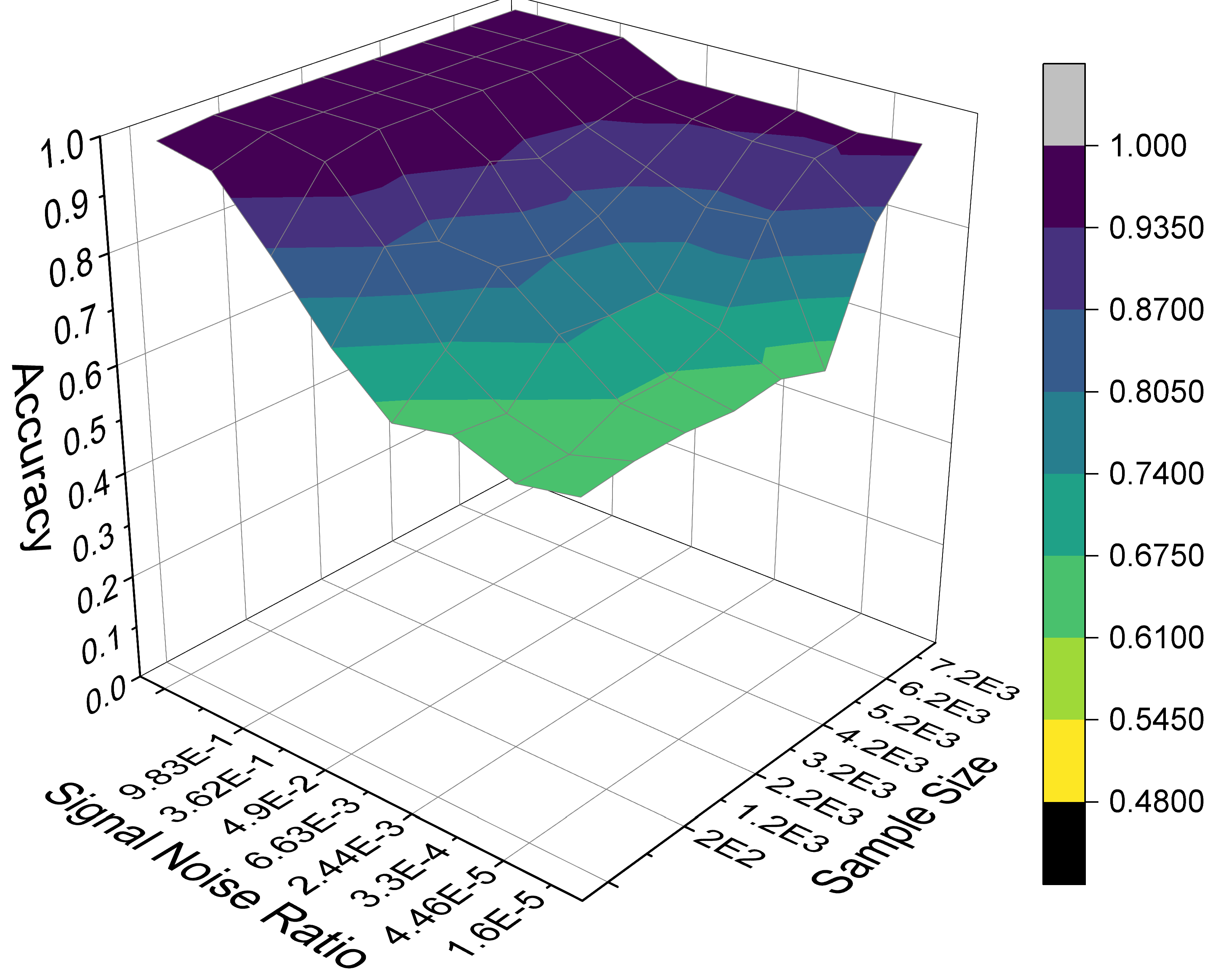}
 }
  \Description{The node classification accuracy under different signal-to-noise ratios and varying sample sizes reflects the degree of GNN's overfitting to noise.}
    \caption{The node classification accuracy under different signal-to-noise ratios.}
    \label{F-experiment-SNR}
\end{figure}

\begin{figure}[!t]
	\centering
   \subfigure[Node Imbalance]{
		\includegraphics[width=0.46\linewidth]{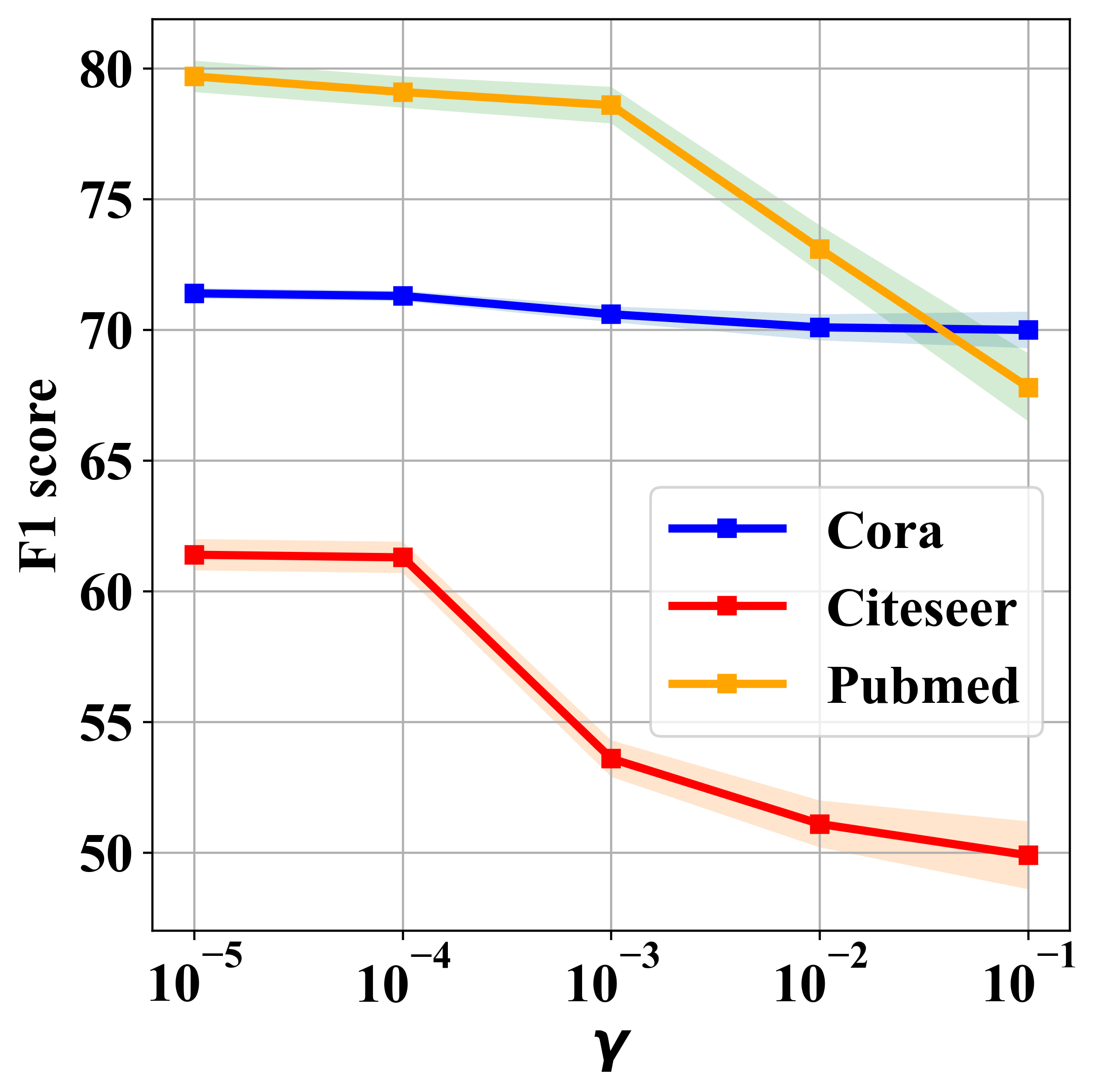}
 }
   \subfigure[OOD Generalization]{
		\includegraphics[width=0.46\linewidth]{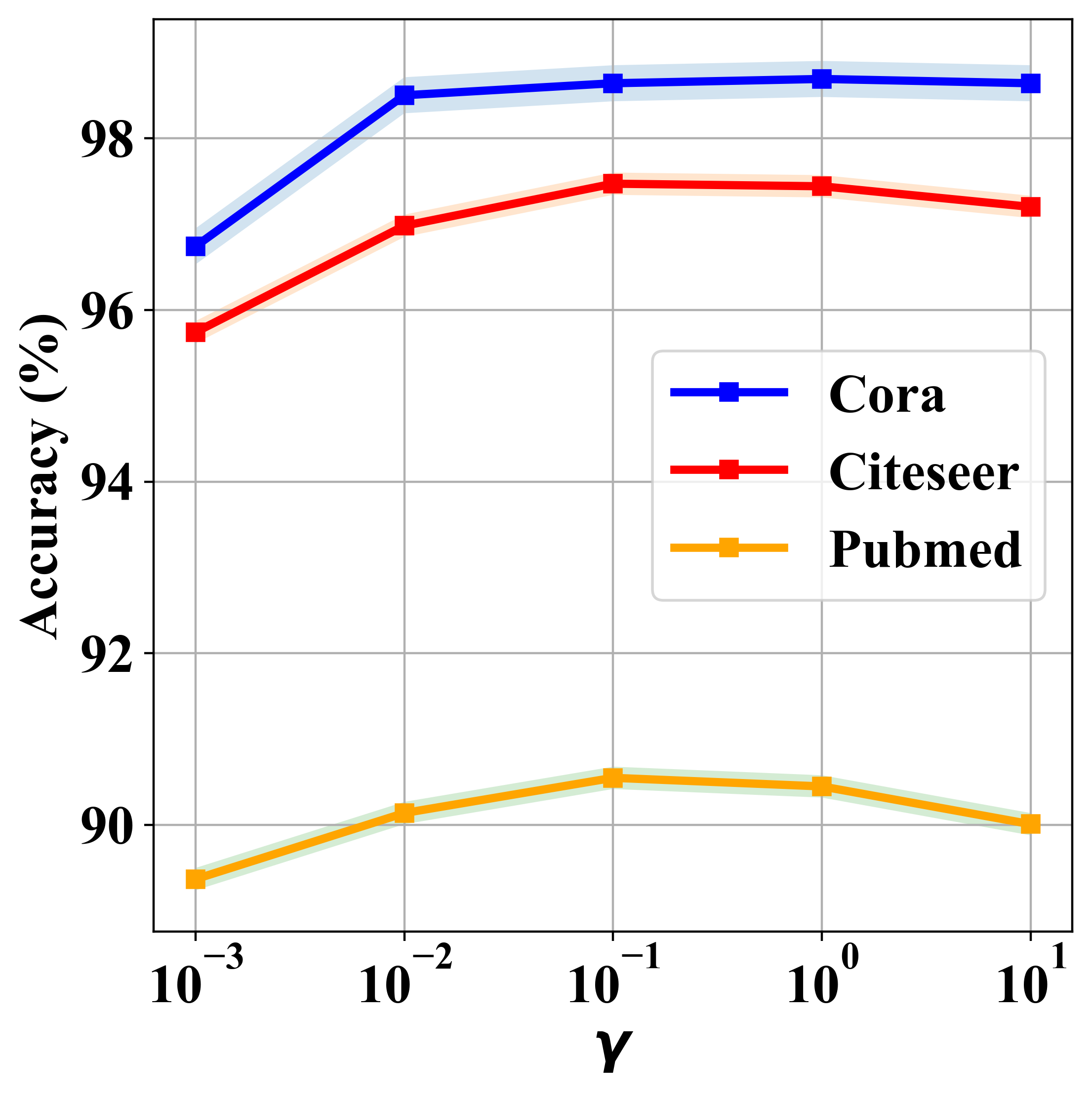}
 }
 \Description{Hyperparameter analysis of $\gamma$ on Cora, Citeseer and Pubmed.}
    \caption{Hyperparameter analysis of $\gamma$ with GCN.}
    \label{F-experiment-hypermeter}
\end{figure}

\textbf{Comparison with $\mathcal{L}_{bound}$.}
$\mathcal{L}_{bound}$ is the method most similar to ours, as it also directly optimizes inter-class consistency of representation norms. However, its main drawback is lacking Lipschitz continuity, leading to unstable optimization and suboptimal results. Figures \ref{F-Visual}(d) and (i) show the visualizations of node representations from the training and test sets using $\mathcal{L}_{bound}$. Some overlap remains in regions with smaller norms, and minority class representations have slightly smaller norms compared to our method (Figure \ref{F-Visual}(j)). Notably, Table \ref{tabel-LOSS-Imbalance} shows that in experiments with imbalanced data, $\mathcal{L}_{bound}$ not only fails to improve performance but degrades the classification ability of GNNs. We believe that imbalanced samples exacerbate the optimization instability of $\mathcal{L}_{bound}$.


\subsection{Reduce Overfitting to Noise. (RQ4)}\label{subSec-experiment-PopulationRisk}



We conduct experimental validation from the perspective of benign overfitting\citep{bartlett2020benign,cao2022benign,huang2023graph} in feature learning to verify whether our method can reduce GNNs overfitting to noise. We adopt the experimental framework from \cite{cao2022benign,huang2023graph},
generating the node feature matrix $\boldsymbol{X} \in \mathbb{R}^{n \times 2d}$ by using a \textit{signal-noise model} (SNM), which involves a Rademacher random variable $y \in \{-1, 1\}$ and a latent vector $\boldsymbol{\mu} \in \mathbb{R}^d$. Specifically, each feature vector is given by $\mathbf{x} = [\mathbf{x}^{(1)}, \mathbf{x}^{(2)}] = [y\boldsymbol{\mu}, \boldsymbol{\xi}]$, where $\mathbf{x}^{(1)}, \mathbf{x}^{(2)} \in \mathbb{R}^d$, and $\boldsymbol{\xi}$ follows a Gaussian distribution $\mathcal{N}(\boldsymbol{0}, \sigma_p^2 \cdot (\boldsymbol{I} - \lVert \boldsymbol{\mu} \rVert_2^{-2} \cdot \boldsymbol{\mu} \boldsymbol{\mu}^{\top}))$, with variance $\sigma_p^2$. We employ a stochastic block model characterized by inter-class edge probability $p$ and intra-class edge probability $s$ to complement this. In this model, the entries of the adjacency matrix $\boldsymbol{A} = (a_{ij})_{n \times n}$ follow a Bernoulli distribution, where $(a_{ij}) \sim \text{Ber}(p)$ if $y_i = y_j$, and $(a_{ij}) \sim \text{Ber}(s)$ if $y_i \neq y_j$. The combined use of the stochastic block model and the signal-noise model is called SNM-SBM($n,p,s,\boldsymbol{\mu},\sigma_p,d$).
We analyze a two-layer GCN denoted as $f$, where the final output is expressed as $f(\boldsymbol{W}, \tilde{\mathbf{x}}) = F_{+1}(\boldsymbol{W}_{+1}, \tilde{\mathbf{x}}) - F_{-1}(\boldsymbol{W}_{-1}, \tilde{\mathbf{x}})$. In this formulation, $F_{+1}(\boldsymbol{W}_{+1}, \tilde{\mathbf{x}})$ and $F_{-1}(\boldsymbol{W}_{-1}, \tilde{\mathbf{x}})$ represent the functions defined as follows:
\begin{equation}
    F_{j}(\boldsymbol{W}_{j},\tilde{\mathbf{x}}) = \frac{1}{m} \sum_{r=1}^{m} \big[ \sigma(\boldsymbol{w}_{j,r}^{\top}\tilde{\mathbf{x}}^{(1)}) + \sigma(\boldsymbol{w}_{j,r}^{\top}\tilde{\mathbf{x}}^{(2)}) \big]
\end{equation}
Here, $\tilde{\boldsymbol{X}} \stackrel{\triangle}{=} [\tilde{\mathbf{x}}_1, \tilde{\mathbf{x}}_2,\cdots,\tilde{\mathbf{x}}_n]^{\top} = \tilde{\boldsymbol{D}}^{-1}\tilde{\boldsymbol{A}}\boldsymbol{X}\in \mathbb{R}^{n \times 2d}$ with $\tilde{\boldsymbol{A}} = \boldsymbol{A} + \boldsymbol{I}_n$ representing the adjacency matrix with self-loop, and $\tilde{\boldsymbol{D}}$ is a diagonal matrix that records the degree of each node.

Given the training dataset $\mathcal{S} \stackrel{\triangle}{=} \{\mathbf{x}_i, y_i \}_{i=1}^n$ and the adjacency matrix $\boldsymbol{A} \in \mathbb{R}^{n \times n}$ generated from the SNM-SBM($n, p, s, \boldsymbol{\mu}, \sigma_p, d$) model, our objective is to optimize the parameter $\boldsymbol{W}$ by minimizing the empirical cross-entropy loss function: $L_{\mathcal{S}}^{GCN}(\boldsymbol{W}) = \frac{1}{n} \sum_{i=1}^n \mathscr{l}_{ce}(y_i \cdot f(\boldsymbol{W, \tilde{\mathbf{x}}_i}))$.  
To evaluate the performance on new test data, which is also generated according to the SNM-SBM distribution, we construct an augmented adjacency matrix $\boldsymbol{A}' \in \mathbb{R}^{(n+1) \times (n+1)}$. This matrix reflects the connections between the new test point and the existing training data points under the stochastic block model. We then analyze the population loss by considering the expectation of the randomness of the new test data point. This population loss is expressed as: $L^{GCN}(\boldsymbol{W}) = \mathbb{E}_{(x,y,\boldsymbol{A}^{\prime}) \sim SNM-SBM}\mathscr{l}_{ce}(y \cdot f(\boldsymbol{W, \tilde{\mathbf{x}}}))$. 

We conduct experiments under different signal-to-noise ratios ($SNR = \lVert \boldsymbol{\mu}\rVert_2/(\sigma_p \sqrt{d})\cdot (n(p+s))^{(q-2)/2q}$) and sample sizes. The results in Figure \ref{F-experiment-SNR} indicate that GNNs exhibit better classification performance in the presence of noise when the representation norms are consistent. We attribute this to our method, which enables GNNs to reduce overfitting to noise, lowering the population risk. Relevant theoretical proofs are provided in the appendix \ref{Appendix-Proof-intra-class-cnsis}.

\subsection{Hyper-parameter Analysis (RQ5)}\label{subSec-Hyper-parameter-Analysis}

In this section, we perform a sensitivity analysis of the hyperparameter $\gamma \in \{10^{-5}, 10^{-4}, 10^{-3}, 10^{-2}, 10^{-1}\}$ in imbalanced node scenarios. We find that larger $\gamma$ values degrade the performance of NodeReg, with the optimal $\gamma$ being less than $10^{-4}$. This decline in performance may be due to an increased variance in norms between the minority and majority classes, where larger $\gamma$ values lead to sharper gradient changes, which hinders model convergence. 

In the OOD scenario, the model exhibits less sensitivity to $\gamma \in \{10^{-3}, 10^{-2}, 10^{-1}, 10^{0}, 10^{1}\}$, although smaller $\gamma$ values slightly reduce performance, with the optimal $\gamma$ around $10^{-1}$. We hypothesize that a smaller $\gamma$ weakens the norm constraint, making it less effective. In conclusion, $\gamma$ plays a more critical role in the imbalanced node scenario than in the OOD generalization scenario, where the model appears more robust to a wider range of $\gamma$ values.

\section{Conclusion}

We identify a novel connection between node imbalance and OOD generalization in semi-supervised node classification tasks linked to norm imbalance of node representations. Based on this insight, we propose a regularization-based optimization method that enforces consistency in the norms of node representations. We conduct extensive empirical studies in both imbalanced and OOD generalization scenarios, demonstrating that our method significantly improves the performance of GNNs in these challenging contexts. Additionally, we provide a theoretical analysis explaining how consistent node representation norms enhance the ability of GNNs to mitigate the effects of imbalance and distributional shifts. We believe that NodeReg has the potential to become a common component for tasks related to semi-supervised node classification.

\textbf{Futrue works.} In future work, we can further explore several interesting aspects of node representation norms. One promising direction is investigating the relationship between node representation norms and noisy labels\citep{dai2021nrgnn,zhu2024robust,cheng2024resurrecting,ding2024divide}. Another potential research avenue is to establish connections between node representation norms and studies on node importance\cite{freeman1977set,park2019estimating}.



\bibliographystyle{ACM-Reference-Format}
\bibliography{sample-base}


\begin{thebibliography}{48}


\ifx \showCODEN    \undefined \def \showCODEN     #1{\unskip}     \fi
\ifx \showDOI      \undefined \def \showDOI       #1{#1}\fi
\ifx \showISBNx    \undefined \def \showISBNx     #1{\unskip}     \fi
\ifx \showISBNxiii \undefined \def \showISBNxiii  #1{\unskip}     \fi
\ifx \showISSN     \undefined \def \showISSN      #1{\unskip}     \fi
\ifx \showLCCN     \undefined \def \showLCCN      #1{\unskip}     \fi
\ifx \shownote     \undefined \def \shownote      #1{#1}          \fi
\ifx \showarticletitle \undefined \def \showarticletitle #1{#1}   \fi
\ifx \showURL      \undefined \def \showURL       {\relax}        \fi
\providecommand\bibfield[2]{#2}
\providecommand\bibinfo[2]{#2}
\providecommand\natexlab[1]{#1}
\providecommand\showeprint[2][]{arXiv:#2}

\bibitem[Arjovsky et~al\mbox{.}(2019)]%
        {arjovsky2019invariant}
\bibfield{author}{\bibinfo{person}{Martin Arjovsky}, \bibinfo{person}{L{\'e}on Bottou}, \bibinfo{person}{Ishaan Gulrajani}, {and} \bibinfo{person}{David Lopez-Paz}.} \bibinfo{year}{2019}\natexlab{}.
\newblock \showarticletitle{Invariant risk minimization}.
\newblock \bibinfo{journal}{\emph{arXiv preprint arXiv:1907.02893}} (\bibinfo{year}{2019}).
\newblock


\bibitem[Baek et~al\mbox{.}(2020)]%
        {baek2020learning}
\bibfield{author}{\bibinfo{person}{Jinheon Baek}, \bibinfo{person}{Dong~Bok Lee}, {and} \bibinfo{person}{Sung~Ju Hwang}.} \bibinfo{year}{2020}\natexlab{}.
\newblock \showarticletitle{Learning to extrapolate knowledge: Transductive few-shot out-of-graph link prediction}.
\newblock \bibinfo{journal}{\emph{Advances in Neural Information Processing Systems}}  \bibinfo{volume}{33} (\bibinfo{year}{2020}), \bibinfo{pages}{546--560}.
\newblock


\bibitem[Bartlett et~al\mbox{.}(2020)]%
        {bartlett2020benign}
\bibfield{author}{\bibinfo{person}{Peter~L Bartlett}, \bibinfo{person}{Philip~M Long}, \bibinfo{person}{G{\'a}bor Lugosi}, {and} \bibinfo{person}{Alexander Tsigler}.} \bibinfo{year}{2020}\natexlab{}.
\newblock \showarticletitle{Benign overfitting in linear regression}.
\newblock \bibinfo{journal}{\emph{Proceedings of the National Academy of Sciences}} \bibinfo{volume}{117}, \bibinfo{number}{48} (\bibinfo{year}{2020}), \bibinfo{pages}{30063--30070}.
\newblock


\bibitem[Ben-Tal et~al\mbox{.}(2013)]%
        {ben2013robust}
\bibfield{author}{\bibinfo{person}{Aharon Ben-Tal}, \bibinfo{person}{Dick Den~Hertog}, \bibinfo{person}{Anja De~Waegenaere}, \bibinfo{person}{Bertrand Melenberg}, {and} \bibinfo{person}{Gijs Rennen}.} \bibinfo{year}{2013}\natexlab{}.
\newblock \showarticletitle{Robust solutions of optimization problems affected by uncertain probabilities}.
\newblock \bibinfo{journal}{\emph{Management Science}} \bibinfo{volume}{59}, \bibinfo{number}{2} (\bibinfo{year}{2013}), \bibinfo{pages}{341--357}.
\newblock


\bibitem[Cao et~al\mbox{.}(2022)]%
        {cao2022benign}
\bibfield{author}{\bibinfo{person}{Yuan Cao}, \bibinfo{person}{Zixiang Chen}, \bibinfo{person}{Misha Belkin}, {and} \bibinfo{person}{Quanquan Gu}.} \bibinfo{year}{2022}\natexlab{}.
\newblock \showarticletitle{Benign overfitting in two-layer convolutional neural networks}.
\newblock \bibinfo{journal}{\emph{Advances in neural information processing systems}}  \bibinfo{volume}{35} (\bibinfo{year}{2022}), \bibinfo{pages}{25237--25250}.
\newblock


\bibitem[Chawla et~al\mbox{.}(2002)]%
        {chawla2002smote}
\bibfield{author}{\bibinfo{person}{Nitesh~V Chawla}, \bibinfo{person}{Kevin~W Bowyer}, \bibinfo{person}{Lawrence~O Hall}, {and} \bibinfo{person}{W~Philip Kegelmeyer}.} \bibinfo{year}{2002}\natexlab{}.
\newblock \showarticletitle{SMOTE: synthetic minority over-sampling technique}.
\newblock \bibinfo{journal}{\emph{Journal of artificial intelligence research}}  \bibinfo{volume}{16} (\bibinfo{year}{2002}), \bibinfo{pages}{321--357}.
\newblock


\bibitem[Chen et~al\mbox{.}(2021)]%
        {chen2021topology}
\bibfield{author}{\bibinfo{person}{Deli Chen}, \bibinfo{person}{Yankai Lin}, \bibinfo{person}{Guangxiang Zhao}, \bibinfo{person}{Xuancheng Ren}, \bibinfo{person}{Peng Li}, \bibinfo{person}{Jie Zhou}, {and} \bibinfo{person}{Xu Sun}.} \bibinfo{year}{2021}\natexlab{}.
\newblock \showarticletitle{Topology-imbalance learning for semi-supervised node classification}.
\newblock \bibinfo{journal}{\emph{Advances in Neural Information Processing Systems}}  \bibinfo{volume}{34} (\bibinfo{year}{2021}), \bibinfo{pages}{29885--29897}.
\newblock


\bibitem[Cheng et~al\mbox{.}(2024)]%
        {cheng2024resurrecting}
\bibfield{author}{\bibinfo{person}{Yao Cheng}, \bibinfo{person}{Caihua Shan}, \bibinfo{person}{Yifei Shen}, \bibinfo{person}{Xiang Li}, \bibinfo{person}{Siqiang Luo}, {and} \bibinfo{person}{Dongsheng Li}.} \bibinfo{year}{2024}\natexlab{}.
\newblock \showarticletitle{Resurrecting Label Propagation for Graphs with Heterophily and Label Noise}. In \bibinfo{booktitle}{\emph{Proceedings of the 30th ACM SIGKDD Conference on Knowledge Discovery and Data Mining}}. \bibinfo{pages}{433--444}.
\newblock


\bibitem[Dai et~al\mbox{.}(2021)]%
        {dai2021nrgnn}
\bibfield{author}{\bibinfo{person}{Enyan Dai}, \bibinfo{person}{Charu Aggarwal}, {and} \bibinfo{person}{Suhang Wang}.} \bibinfo{year}{2021}\natexlab{}.
\newblock \showarticletitle{Nrgnn: Learning a label noise resistant graph neural network on sparsely and noisily labeled graphs}. In \bibinfo{booktitle}{\emph{Proceedings of the 27th ACM SIGKDD conference on knowledge discovery \& data mining}}. \bibinfo{pages}{227--236}.
\newblock


\bibitem[Ding et~al\mbox{.}(2024)]%
        {ding2024divide}
\bibfield{author}{\bibinfo{person}{Kaize Ding}, \bibinfo{person}{Xiaoxiao Ma}, \bibinfo{person}{Yixin Liu}, {and} \bibinfo{person}{Shirui Pan}.} \bibinfo{year}{2024}\natexlab{}.
\newblock \showarticletitle{Divide and Denoise: Empowering Simple Models for Robust Semi-Supervised Node Classification against Label Noise}. In \bibinfo{booktitle}{\emph{Proceedings of the 30th ACM SIGKDD Conference on Knowledge Discovery and Data Mining}}. \bibinfo{pages}{574--584}.
\newblock


\bibitem[Fan et~al\mbox{.}(2019)]%
        {fan2019graph}
\bibfield{author}{\bibinfo{person}{Wenqi Fan}, \bibinfo{person}{Yao Ma}, \bibinfo{person}{Qing Li}, \bibinfo{person}{Yuan He}, \bibinfo{person}{Eric Zhao}, \bibinfo{person}{Jiliang Tang}, {and} \bibinfo{person}{Dawei Yin}.} \bibinfo{year}{2019}\natexlab{}.
\newblock \showarticletitle{Graph neural networks for social recommendation}. In \bibinfo{booktitle}{\emph{The world wide web conference}}. \bibinfo{pages}{417--426}.
\newblock


\bibitem[Fisher(1936)]%
        {fisher1936use}
\bibfield{author}{\bibinfo{person}{Ronald~A Fisher}.} \bibinfo{year}{1936}\natexlab{}.
\newblock \showarticletitle{The use of multiple measurements in taxonomic problems}.
\newblock \bibinfo{journal}{\emph{Annals of eugenics}} \bibinfo{volume}{7}, \bibinfo{number}{2} (\bibinfo{year}{1936}), \bibinfo{pages}{179--188}.
\newblock


\bibitem[Freeman(1977)]%
        {freeman1977set}
\bibfield{author}{\bibinfo{person}{LC Freeman}.} \bibinfo{year}{1977}\natexlab{}.
\newblock \showarticletitle{A set of measures of centrality based on betweenness}.
\newblock \bibinfo{journal}{\emph{Sociometry}} (\bibinfo{year}{1977}).
\newblock


\bibitem[Ganin et~al\mbox{.}(2016)]%
        {ganin2016domain}
\bibfield{author}{\bibinfo{person}{Yaroslav Ganin}, \bibinfo{person}{Evgeniya Ustinova}, \bibinfo{person}{Hana Ajakan}, \bibinfo{person}{Pascal Germain}, \bibinfo{person}{Hugo Larochelle}, \bibinfo{person}{Fran{\c{c}}ois Laviolette}, \bibinfo{person}{Mario March}, {and} \bibinfo{person}{Victor Lempitsky}.} \bibinfo{year}{2016}\natexlab{}.
\newblock \showarticletitle{Domain-adversarial training of neural networks}.
\newblock \bibinfo{journal}{\emph{Journal of machine learning research}} \bibinfo{volume}{17}, \bibinfo{number}{59} (\bibinfo{year}{2016}), \bibinfo{pages}{1--35}.
\newblock


\bibitem[Girshick(2015)]%
        {girshick2015fast}
\bibfield{author}{\bibinfo{person}{R Girshick}.} \bibinfo{year}{2015}\natexlab{}.
\newblock \showarticletitle{Fast r-cnn}.
\newblock \bibinfo{journal}{\emph{arXiv preprint arXiv:1504.08083}} (\bibinfo{year}{2015}).
\newblock


\bibitem[Hamilton et~al\mbox{.}(2017)]%
        {hamilton2017inductive}
\bibfield{author}{\bibinfo{person}{William~L. Hamilton}, \bibinfo{person}{Zhitao Ying}, {and} \bibinfo{person}{Jure Leskovec}.} \bibinfo{year}{2017}\natexlab{}.
\newblock \showarticletitle{Inductive Representation Learning on Large Graphs}. In \bibinfo{booktitle}{\emph{Advances in Neural Information Processing Systems 30: Annual Conference on Neural Information Processing Systems 2017, December 4-9, 2017, Long Beach, CA, {USA}}}, \bibfield{editor}{\bibinfo{person}{Isabelle Guyon}, \bibinfo{person}{Ulrike von Luxburg}, \bibinfo{person}{Samy Bengio}, \bibinfo{person}{Hanna~M. Wallach}, \bibinfo{person}{Rob Fergus}, \bibinfo{person}{S.~V.~N. Vishwanathan}, {and} \bibinfo{person}{Roman Garnett}} (Eds.). \bibinfo{pages}{1024--1034}.
\newblock
\urldef\tempurl%
\url{https://proceedings.neurips.cc/paper/2017/hash/5dd9db5e033da9c6fb5ba83c7a7ebea9-Abstract.html}
\showURL{%
\tempurl}


\bibitem[Han et~al\mbox{.}(2021)]%
        {han2021neural}
\bibfield{author}{\bibinfo{person}{XY Han}, \bibinfo{person}{Vardan Papyan}, {and} \bibinfo{person}{David~L Donoho}.} \bibinfo{year}{2021}\natexlab{}.
\newblock \showarticletitle{Neural collapse under mse loss: Proximity to and dynamics on the central path}.
\newblock \bibinfo{journal}{\emph{arXiv preprint arXiv:2106.02073}} (\bibinfo{year}{2021}).
\newblock


\bibitem[Hu et~al\mbox{.}(2020)]%
        {hu2020open}
\bibfield{author}{\bibinfo{person}{Weihua Hu}, \bibinfo{person}{Matthias Fey}, \bibinfo{person}{Marinka Zitnik}, \bibinfo{person}{Yuxiao Dong}, \bibinfo{person}{Hongyu Ren}, \bibinfo{person}{Bowen Liu}, \bibinfo{person}{Michele Catasta}, {and} \bibinfo{person}{Jure Leskovec}.} \bibinfo{year}{2020}\natexlab{}.
\newblock \showarticletitle{Open graph benchmark: Datasets for machine learning on graphs}.
\newblock \bibinfo{journal}{\emph{Advances in neural information processing systems}}  \bibinfo{volume}{33} (\bibinfo{year}{2020}), \bibinfo{pages}{22118--22133}.
\newblock


\bibitem[Huang et~al\mbox{.}(2023)]%
        {huang2023graph}
\bibfield{author}{\bibinfo{person}{Wei Huang}, \bibinfo{person}{Yuan Cao}, \bibinfo{person}{Haonan Wang}, \bibinfo{person}{Xin Cao}, {and} \bibinfo{person}{Taiji Suzuki}.} \bibinfo{year}{2023}\natexlab{}.
\newblock \showarticletitle{Graph neural networks provably benefit from structural information: A feature learning perspective}.
\newblock \bibinfo{journal}{\emph{arXiv preprint arXiv:2306.13926}} (\bibinfo{year}{2023}).
\newblock


\bibitem[Khosla et~al\mbox{.}(2020)]%
        {Khosla_NIPS20_SupCon}
\bibfield{author}{\bibinfo{person}{Prannay Khosla}, \bibinfo{person}{Piotr Teterwak}, \bibinfo{person}{Chen Wang}, \bibinfo{person}{Aaron Sarna}, \bibinfo{person}{Yonglong Tian}, \bibinfo{person}{Phillip Isola}, \bibinfo{person}{Aaron Maschinot}, \bibinfo{person}{Ce Liu}, {and} \bibinfo{person}{Dilip Krishnan}.} \bibinfo{year}{2020}\natexlab{}.
\newblock \showarticletitle{Supervised Contrastive Learning}. In \bibinfo{booktitle}{\emph{Advances in Neural Information Processing Systems 33: Annual Conference on Neural Information Processing Systems 2020, NeurIPS 2020, December 6-12, 2020, virtual}}, \bibfield{editor}{\bibinfo{person}{Hugo Larochelle}, \bibinfo{person}{Marc'Aurelio Ranzato}, \bibinfo{person}{Raia Hadsell}, \bibinfo{person}{Maria{-}Florina Balcan}, {and} \bibinfo{person}{Hsuan{-}Tien Lin}} (Eds.).
\newblock
\urldef\tempurl%
\url{https://proceedings.neurips.cc/paper/2020/hash/d89a66c7c80a29b1bdbab0f2a1a94af8-Abstract.html}
\showURL{%
\tempurl}


\bibitem[Kipf and Welling(2016)]%
        {kipf2016semi}
\bibfield{author}{\bibinfo{person}{Thomas~N Kipf} {and} \bibinfo{person}{Max Welling}.} \bibinfo{year}{2016}\natexlab{}.
\newblock \showarticletitle{Semi-supervised classification with graph convolutional networks}.
\newblock \bibinfo{journal}{\emph{arXiv preprint arXiv:1609.02907}} (\bibinfo{year}{2016}).
\newblock


\bibitem[Namkoong and Duchi(2016)]%
        {namkoong2016stochastic}
\bibfield{author}{\bibinfo{person}{Hongseok Namkoong} {and} \bibinfo{person}{John~C Duchi}.} \bibinfo{year}{2016}\natexlab{}.
\newblock \showarticletitle{Stochastic gradient methods for distributionally robust optimization with f-divergences}.
\newblock \bibinfo{journal}{\emph{Advances in neural information processing systems}}  \bibinfo{volume}{29} (\bibinfo{year}{2016}).
\newblock


\bibitem[Papyan et~al\mbox{.}(2020)]%
        {papyan2020prevalence}
\bibfield{author}{\bibinfo{person}{Vardan Papyan}, \bibinfo{person}{XY Han}, {and} \bibinfo{person}{David~L Donoho}.} \bibinfo{year}{2020}\natexlab{}.
\newblock \showarticletitle{Prevalence of neural collapse during the terminal phase of deep learning training}.
\newblock \bibinfo{journal}{\emph{Proceedings of the National Academy of Sciences}} \bibinfo{volume}{117}, \bibinfo{number}{40} (\bibinfo{year}{2020}), \bibinfo{pages}{24652--24663}.
\newblock


\bibitem[Park et~al\mbox{.}(2021)]%
        {park2021graphens}
\bibfield{author}{\bibinfo{person}{Joonhyung Park}, \bibinfo{person}{Jaeyun Song}, {and} \bibinfo{person}{Eunho Yang}.} \bibinfo{year}{2021}\natexlab{}.
\newblock \showarticletitle{Graphens: Neighbor-aware ego network synthesis for class-imbalanced node classification}. In \bibinfo{booktitle}{\emph{International conference on learning representations}}.
\newblock


\bibitem[Park et~al\mbox{.}(2019)]%
        {park2019estimating}
\bibfield{author}{\bibinfo{person}{Namyong Park}, \bibinfo{person}{Andrey Kan}, \bibinfo{person}{Xin~Luna Dong}, \bibinfo{person}{Tong Zhao}, {and} \bibinfo{person}{Christos Faloutsos}.} \bibinfo{year}{2019}\natexlab{}.
\newblock \showarticletitle{Estimating node importance in knowledge graphs using graph neural networks}. In \bibinfo{booktitle}{\emph{Proceedings of the 25th ACM SIGKDD international conference on knowledge discovery \& data mining}}. \bibinfo{pages}{596--606}.
\newblock


\bibitem[Ren et~al\mbox{.}(2018)]%
        {ren2018learning}
\bibfield{author}{\bibinfo{person}{Mengye Ren}, \bibinfo{person}{Wenyuan Zeng}, \bibinfo{person}{Bin Yang}, {and} \bibinfo{person}{Raquel Urtasun}.} \bibinfo{year}{2018}\natexlab{}.
\newblock \showarticletitle{Learning to reweight examples for robust deep learning}. In \bibinfo{booktitle}{\emph{International conference on machine learning}}. PMLR, \bibinfo{pages}{4334--4343}.
\newblock


\bibitem[Rozemberczki and Sarkar(2021)]%
        {rozemberczki2021twitch}
\bibfield{author}{\bibinfo{person}{Benedek Rozemberczki} {and} \bibinfo{person}{Rik Sarkar}.} \bibinfo{year}{2021}\natexlab{}.
\newblock \showarticletitle{Twitch gamers: a dataset for evaluating proximity preserving and structural role-based node embeddings}.
\newblock \bibinfo{journal}{\emph{arXiv preprint arXiv:2101.03091}} (\bibinfo{year}{2021}).
\newblock


\bibitem[Sagawa et~al\mbox{.}(2019)]%
        {sagawa2019distributionally}
\bibfield{author}{\bibinfo{person}{Shiori Sagawa}, \bibinfo{person}{Pang~Wei Koh}, \bibinfo{person}{Tatsunori~B Hashimoto}, {and} \bibinfo{person}{Percy Liang}.} \bibinfo{year}{2019}\natexlab{}.
\newblock \showarticletitle{Distributionally robust neural networks for group shifts: On the importance of regularization for worst-case generalization}.
\newblock \bibinfo{journal}{\emph{arXiv preprint arXiv:1911.08731}} (\bibinfo{year}{2019}).
\newblock


\bibitem[Shchur et~al\mbox{.}(2018)]%
        {shchur2018pitfalls}
\bibfield{author}{\bibinfo{person}{Oleksandr Shchur}, \bibinfo{person}{Maximilian Mumme}, \bibinfo{person}{Aleksandar Bojchevski}, {and} \bibinfo{person}{Stephan G{\"u}nnemann}.} \bibinfo{year}{2018}\natexlab{}.
\newblock \showarticletitle{Pitfalls of graph neural network evaluation}.
\newblock \bibinfo{journal}{\emph{arXiv preprint arXiv:1811.05868}} (\bibinfo{year}{2018}).
\newblock


\bibitem[Shi et~al\mbox{.}(2020)]%
        {shi2020multi}
\bibfield{author}{\bibinfo{person}{Min Shi}, \bibinfo{person}{Yufei Tang}, \bibinfo{person}{Xingquan Zhu}, \bibinfo{person}{David Wilson}, {and} \bibinfo{person}{Jianxun Liu}.} \bibinfo{year}{2020}\natexlab{}.
\newblock \showarticletitle{Multi-class imbalanced graph convolutional network learning}. In \bibinfo{booktitle}{\emph{Proceedings of the Twenty-Ninth International Joint Conference on Artificial Intelligence (IJCAI-20)}}.
\newblock


\bibitem[Sun et~al\mbox{.}(2016)]%
        {sun2016return}
\bibfield{author}{\bibinfo{person}{Baochen Sun}, \bibinfo{person}{Jiashi Feng}, {and} \bibinfo{person}{Kate Saenko}.} \bibinfo{year}{2016}\natexlab{}.
\newblock \showarticletitle{Return of frustratingly easy domain adaptation}. In \bibinfo{booktitle}{\emph{Proceedings of the AAAI conference on artificial intelligence}}, Vol.~\bibinfo{volume}{30}.
\newblock


\bibitem[Sun and Saenko(2016)]%
        {sun2016deep}
\bibfield{author}{\bibinfo{person}{Baochen Sun} {and} \bibinfo{person}{Kate Saenko}.} \bibinfo{year}{2016}\natexlab{}.
\newblock \showarticletitle{Deep coral: Correlation alignment for deep domain adaptation}. In \bibinfo{booktitle}{\emph{Computer Vision--ECCV 2016 Workshops: Amsterdam, The Netherlands, October 8-10 and 15-16, 2016, Proceedings, Part III 14}}. Springer, \bibinfo{pages}{443--450}.
\newblock


\bibitem[Tirer and Bruna(2022)]%
        {tirer2022extended}
\bibfield{author}{\bibinfo{person}{Tom Tirer} {and} \bibinfo{person}{Joan Bruna}.} \bibinfo{year}{2022}\natexlab{}.
\newblock \showarticletitle{Extended unconstrained features model for exploring deep neural collapse}. In \bibinfo{booktitle}{\emph{International Conference on Machine Learning}}. PMLR, \bibinfo{pages}{21478--21505}.
\newblock


\bibitem[Velickovic et~al\mbox{.}(2018)]%
        {2017Graph}
\bibfield{author}{\bibinfo{person}{Petar Velickovic}, \bibinfo{person}{Guillem Cucurull}, \bibinfo{person}{Arantxa Casanova}, \bibinfo{person}{Adriana Romero}, \bibinfo{person}{Pietro Li{\`{o}}}, {and} \bibinfo{person}{Yoshua Bengio}.} \bibinfo{year}{2018}\natexlab{}.
\newblock \showarticletitle{Graph Attention Networks}. In \bibinfo{booktitle}{\emph{6th International Conference on Learning Representations, {ICLR} 2018, Vancouver, BC, Canada, April 30 - May 3, 2018, Conference Track Proceedings}}. \bibinfo{publisher}{OpenReview.net}.
\newblock
\urldef\tempurl%
\url{https://openreview.net/forum?id=rJXMpikCZ}
\showURL{%
\tempurl}


\bibitem[Wang et~al\mbox{.}(2019)]%
        {wang2019knowledge}
\bibfield{author}{\bibinfo{person}{Hongwei Wang}, \bibinfo{person}{Miao Zhao}, \bibinfo{person}{Xing Xie}, \bibinfo{person}{Wenjie Li}, {and} \bibinfo{person}{Minyi Guo}.} \bibinfo{year}{2019}\natexlab{}.
\newblock \showarticletitle{Knowledge graph convolutional networks for recommender systems}. In \bibinfo{booktitle}{\emph{The world wide web conference}}. \bibinfo{pages}{3307--3313}.
\newblock


\bibitem[Wen et~al\mbox{.}(2016)]%
        {wen2016discriminative}
\bibfield{author}{\bibinfo{person}{Yandong Wen}, \bibinfo{person}{Kaipeng Zhang}, \bibinfo{person}{Zhifeng Li}, {and} \bibinfo{person}{Yu Qiao}.} \bibinfo{year}{2016}\natexlab{}.
\newblock \showarticletitle{A discriminative feature learning approach for deep face recognition}. In \bibinfo{booktitle}{\emph{Computer vision--ECCV 2016: 14th European conference, amsterdam, the netherlands, October 11--14, 2016, proceedings, part VII 14}}. Springer, \bibinfo{pages}{499--515}.
\newblock


\bibitem[Wu et~al\mbox{.}(2024)]%
        {wu2024graph}
\bibfield{author}{\bibinfo{person}{Qitian Wu}, \bibinfo{person}{Fan Nie}, \bibinfo{person}{Chenxiao Yang}, \bibinfo{person}{Tianyi Bao}, {and} \bibinfo{person}{Junchi Yan}.} \bibinfo{year}{2024}\natexlab{}.
\newblock \showarticletitle{Graph out-of-distribution generalization via causal intervention}. In \bibinfo{booktitle}{\emph{Proceedings of the ACM on Web Conference 2024}}. \bibinfo{pages}{850--860}.
\newblock


\bibitem[Wu et~al\mbox{.}(2022)]%
        {wu2022handling}
\bibfield{author}{\bibinfo{person}{Qitian Wu}, \bibinfo{person}{Hengrui Zhang}, \bibinfo{person}{Junchi Yan}, {and} \bibinfo{person}{David Wipf}.} \bibinfo{year}{2022}\natexlab{}.
\newblock \showarticletitle{Handling distribution shifts on graphs: An invariance perspective}.
\newblock \bibinfo{journal}{\emph{arXiv preprint arXiv:2202.02466}} (\bibinfo{year}{2022}).
\newblock


\bibitem[Yan et~al\mbox{.}(2024)]%
        {yan2024rethinking}
\bibfield{author}{\bibinfo{person}{Divin Yan}, \bibinfo{person}{Gengchen Wei}, \bibinfo{person}{Chen Yang}, \bibinfo{person}{Shengzhong Zhang}, {et~al\mbox{.}}} \bibinfo{year}{2024}\natexlab{}.
\newblock \showarticletitle{Rethinking semi-supervised imbalanced node classification from bias-variance decomposition}.
\newblock \bibinfo{journal}{\emph{Advances in Neural Information Processing Systems}}  \bibinfo{volume}{36} (\bibinfo{year}{2024}).
\newblock


\bibitem[Yang et~al\mbox{.}({[n.\,d.]})]%
        {yangbounded}
\bibfield{author}{\bibinfo{person}{Shenzhi Yang}, \bibinfo{person}{Bin Liang}, \bibinfo{person}{An Liu}, \bibinfo{person}{Lin Gui}, \bibinfo{person}{Xingkai Yao}, {and} \bibinfo{person}{Xiaofang Zhang}.} \bibinfo{year}{[n.\,d.]}\natexlab{}.
\newblock \showarticletitle{Bounded and Uniform Energy-based Out-of-distribution Detection for Graphs}. In \bibinfo{booktitle}{\emph{Forty-first International Conference on Machine Learning}}.
\newblock


\bibitem[Zhang(2017)]%
        {zhang2017mixup}
\bibfield{author}{\bibinfo{person}{Hongyi Zhang}.} \bibinfo{year}{2017}\natexlab{}.
\newblock \showarticletitle{mixup: Beyond empirical risk minimization}.
\newblock \bibinfo{journal}{\emph{arXiv preprint arXiv:1710.09412}} (\bibinfo{year}{2017}).
\newblock


\bibitem[Zhang and Chen(2019)]%
        {zhang2019inductive}
\bibfield{author}{\bibinfo{person}{Muhan Zhang} {and} \bibinfo{person}{Yixin Chen}.} \bibinfo{year}{2019}\natexlab{}.
\newblock \showarticletitle{Inductive matrix completion based on graph neural networks}.
\newblock \bibinfo{journal}{\emph{arXiv preprint arXiv:1904.12058}} (\bibinfo{year}{2019}).
\newblock


\bibitem[Zhang et~al\mbox{.}(2024)]%
        {zhang2024bim}
\bibfield{author}{\bibinfo{person}{Wentao Zhang}, \bibinfo{person}{Xinyi Gao}, \bibinfo{person}{Ling Yang}, \bibinfo{person}{Meng Cao}, \bibinfo{person}{Ping Huang}, \bibinfo{person}{Jiulong Shan}, \bibinfo{person}{Hongzhi Yin}, {and} \bibinfo{person}{Bin Cui}.} \bibinfo{year}{2024}\natexlab{}.
\newblock \showarticletitle{BIM: Improving Graph Neural Networks with Balanced Influence Maximization}. In \bibinfo{booktitle}{\emph{2024 IEEE 40th International Conference on Data Engineering (ICDE)}}. IEEE, \bibinfo{pages}{2931--2944}.
\newblock


\bibitem[Zhao et~al\mbox{.}(2021)]%
        {zhao2021graphsmote}
\bibfield{author}{\bibinfo{person}{Tianxiang Zhao}, \bibinfo{person}{Xiang Zhang}, {and} \bibinfo{person}{Suhang Wang}.} \bibinfo{year}{2021}\natexlab{}.
\newblock \showarticletitle{Graphsmote: Imbalanced node classification on graphs with graph neural networks}. In \bibinfo{booktitle}{\emph{Proceedings of the 14th ACM international conference on web search and data mining}}. \bibinfo{pages}{833--841}.
\newblock


\bibitem[Zhou et~al\mbox{.}(2022)]%
        {zhou2022all}
\bibfield{author}{\bibinfo{person}{Jinxin Zhou}, \bibinfo{person}{Chong You}, \bibinfo{person}{Xiao Li}, \bibinfo{person}{Kangning Liu}, \bibinfo{person}{Sheng Liu}, \bibinfo{person}{Qing Qu}, {and} \bibinfo{person}{Zhihui Zhu}.} \bibinfo{year}{2022}\natexlab{}.
\newblock \showarticletitle{Are all losses created equal: A neural collapse perspective}.
\newblock \bibinfo{journal}{\emph{Advances in Neural Information Processing Systems}}  \bibinfo{volume}{35} (\bibinfo{year}{2022}), \bibinfo{pages}{31697--31710}.
\newblock


\bibitem[Zhu et~al\mbox{.}(2021b)]%
        {zhu2021shift}
\bibfield{author}{\bibinfo{person}{Qi Zhu}, \bibinfo{person}{Natalia Ponomareva}, \bibinfo{person}{Jiawei Han}, {and} \bibinfo{person}{Bryan Perozzi}.} \bibinfo{year}{2021}\natexlab{b}.
\newblock \showarticletitle{Shift-robust gnns: Overcoming the limitations of localized graph training data}.
\newblock \bibinfo{journal}{\emph{Advances in Neural Information Processing Systems}}  \bibinfo{volume}{34} (\bibinfo{year}{2021}), \bibinfo{pages}{27965--27977}.
\newblock


\bibitem[Zhu et~al\mbox{.}(2024)]%
        {zhu2024robust}
\bibfield{author}{\bibinfo{person}{Yonghua Zhu}, \bibinfo{person}{Lei Feng}, \bibinfo{person}{Zhenyun Deng}, \bibinfo{person}{Yang Chen}, \bibinfo{person}{Robert Amor}, {and} \bibinfo{person}{Michael Witbrock}.} \bibinfo{year}{2024}\natexlab{}.
\newblock \showarticletitle{Robust Node Classification on Graph Data with Graph and Label Noise}. In \bibinfo{booktitle}{\emph{Proceedings of the AAAI Conference on Artificial Intelligence}}, Vol.~\bibinfo{volume}{38}. \bibinfo{pages}{17220--17227}.
\newblock


\bibitem[Zhu et~al\mbox{.}(2021a)]%
        {zhu2021geometric}
\bibfield{author}{\bibinfo{person}{Zhihui Zhu}, \bibinfo{person}{Tianyu Ding}, \bibinfo{person}{Jinxin Zhou}, \bibinfo{person}{Xiao Li}, \bibinfo{person}{Chong You}, \bibinfo{person}{Jeremias Sulam}, {and} \bibinfo{person}{Qing Qu}.} \bibinfo{year}{2021}\natexlab{a}.
\newblock \showarticletitle{A geometric analysis of neural collapse with unconstrained features}.
\newblock \bibinfo{journal}{\emph{Advances in Neural Information Processing Systems}}  \bibinfo{volume}{34} (\bibinfo{year}{2021}), \bibinfo{pages}{29820--29834}.
\newblock


\end{thebibliography}


\appendix
\section{Experiment Settings}\label{Sec-Appendix-Exp-Setting}
\subsection{Node Imbalance}


\paragraph{\textbf{Baseline Methods.}} We compare several representative methods for addressing class imbalance, which include:
\begin{itemize}[nosep, topsep=0pt, leftmargin=*]
    \item \textbf{Random Over-Sampling (ROS)}: Re-samples nodes and their edges in minority classes to balance class distribution.
    \item \textbf{SMOTE}\citep{chawla2002smote}: Generates synthetic samples by interpolating between a minority sample and its nearest neighbors within the same class. The synthetic node retains the same edges as the target node.
    \item \textbf{Reweight}\citep{ren2018learning}: A cost-sensitive approach that adjusts the classification loss by increasing the weight assigned to minority classes.
    \item \textbf{DR-GCN}\citep{shi2020multi}: Enhances representation learning using conditional adversarial training and distribution alignment to better separate nodes from different classes.
    \item \textbf{GraphSMOTE}\citep{zhao2021graphsmote}: Synthesizes new minority nodes in the graph's embedding space and generates edges by training an edge generator.
    \item \textbf{ReNode}\citep{chen2021topology}: Addresses graph topology imbalance by re-weighting the influence of labeled nodes based on their relative positions to class boundaries.
    \item \textbf{GraphENS}\citep{park2021graphens}: Constructs synthetic ego networks by comparing source ego networks while preventing the introduction of harmful features during node generation by leveraging node feature saliency.
    \item \textbf{ReVar}\citep{yan2024rethinking}: Uses graph augmentation to estimate variance and mitigates the effects of imbalance and model variance by introducing regularization across different views.
    \item \textbf{BIM}\citep{zhang2024bim} suggests that node imbalance is influenced not only by class sample imbalance but also by an imbalance in the receptive field of nodes. 
\end{itemize}

\subsection{OOD Generalization}

\paragraph{\textbf{Baseline Methods.}} We compare several representative methods for addressing distribution shift, which include:
\begin{itemize}[nosep, topsep=0pt, leftmargin=*]
\item \textbf{Empirical risk minimization (ERM)} trains the model with standard supervised loss. 
\item \textbf{Invariant risk minimization (IRM)}\citep{arjovsky2019invariant} is a learning paradigm to estimate invariant correlations across multiple training distributions. 

\item \textbf{Deep CORAL}\citep{sun2016deep} extends CORAL\citep{sun2016return} to learn a nonlinear transformation that aligns the correlation of layer activations. 

\item \textbf{DANN}\citep{ganin2016domain} introduce a new method for learning domain-adapted representations, in which the data at training and testing come from similar but different distributions. 

\item \textbf{GroupDRO}\citep{sagawa2019distributionally} shows that regularisation is essential for worst group generalization in over-parameterized regimes, even if it is not for average generalization. By combining the group DRO model\citep{ben2013robust,namkoong2016stochastic} with increased regularisation, over the typical L2 regularisation or early stopping, GroupDRO achieves higher worst group accuracy.

\item \textbf{Mixup}\citep{zhang2017mixup} aims to increase training data by interpolating between observed samples. 

\item \textbf{SR-GNN}\citep{zhu2021shift} is designed to explain the distributional discrepancy between biased training data and the accurate inference distribution over graphs by adjusting the GNN model to account for the distributional shift between labeled nodes in the training set and the rest of the dataset. 

\item \textbf{EERM}\citep{wu2022handling} utilizes the invariance principle to develop an adversarial training method for environmental exploration.

\item \textbf{CaNet}\citep{wu2024graph} enhances the OOD generalization capability of GNNs by preserving causal invariance among nodes.
\end{itemize}
We use GCN and GAT as their encoder backbones for all the competitors, respectively.



\section{Theory Proof}



\subsection{Proof of Property\ref{property-lipschitz}} \label{proof-lp1}

\begin{proof}
Consider the first-order derivative of $\mathcal{L}_{\mathrm{NodeReg}}^v$:
    \begin{equation}\label{equa-proof-L_{NodeReg-3}-g1}
    \frac{d}{d\delta}\mathcal{L}_{\mathrm{NodeReg}}^v = 
    \begin{cases} 
        1 & \text{if } \delta_v \geq \gamma, \\
        \frac{\delta}{\gamma} & \text{if } |\delta_v| \leq \gamma\\
       -1 & \text{if }  \delta_v \leq -\gamma
    \end{cases}
\end{equation}
We find that the first-order derivative of \( \mathcal{L}_{\mathrm{NodeReg}}^v \) has a maximum value of 1. Thus, the Property 4.1 is proofed.

\end{proof}

\subsection{Proof of Property \ref{property-lipschitz gradient}}\label{proof-lp2}

Consider the second-order derivative of $\mathcal{L}_{\mathrm{NodeReg}}^v$:
    \begin{equation}\label{equa-proof-L_{NodeReg-3}-g2}
    \frac{d^{2}}{d\delta^{2}}\mathcal{L}_{\mathrm{NodeReg}}^v = 
    \begin{cases} 
        0 & \text{if } \delta_v \geq \gamma, \\
        \frac{1}{\gamma} & \text{if } |\delta_v| \leq \gamma\\
        0 & \text{if } \delta \leq -\gamma
    \end{cases}
\end{equation}
We find that the second-order derivative of \( \mathcal{L}_{\mathrm{NodeReg}}^v \) has a maximum value of $\frac{1}{\gamma}$. Thus, the Property 4.2 is proofed.



\subsection{Proof of Proposition \ref{proposition-intra-class}}\label{Appendix-Proof-intra-class-cnsis}

\begin{lemma}\label{lemma-1}\citep{huang2023graph}
With probability, at least 
\begin{equation}
    1-4mT \cdot exp(-C_2^{-1}\sigma_0^{2}\sigma_p^{2}d^{-1}n(p+s))
\end{equation}
, we have that $max_{j,r}|\langle \boldsymbol{w}_{j,r}^{(t), \tilde{\boldsymbol{\xi}}}\rangle| \leq 1/2 $ for all $0\leq t \leq T$, where $C_2 = \tilde{O}(1)$.
\end{lemma}

Consider the occurrence of event $\mathscr{E}$, defined as the condition under which \ref{lemma-1} is satisfied. We can then express the population loss $L^{GCN}(\boldsymbol{W^{(t)}})$ as a sum of two components:

\begin{align}
         \mathbb{E}[\mathscr{l}_{ce}(if(\boldsymbol{W^{(t)}}, \tilde{\mathbf{x}}))] &=\underbrace{\mathbb{E}[\mathbbm{1}(\mathscr{E})\mathscr{l}_{ce}(yf(\boldsymbol{W^{(t)}}, \tilde{\mathbf{x}}))]}_{Term \ I_1 \ (Empirical \ Risk)}\\
         &+ \underbrace{\mathbb{E}[\mathbbm{1}(\mathscr{E}^c)\mathscr{l}_{ce}(yf(\boldsymbol{W^{(t)}}, \tilde{\mathbf{x}}))]}_{Term \ I_2 \ (Generalization \ Gap)}
\end{align}

\textbf{Estimating $I_2$}:By selecting an arbitrary training data point $(\mathbf{x}_{i^{\prime}}, y_{i^{\prime}})$ with $y_{i^{\prime}} = y$, We can derive the following:
\begin{equation}
    I_2 \leq \sqrt{\mathbb{E}[\mathbbm{1}(\mathscr{E}^c)]} \cdot \sqrt{\mathbb{E}\big[   \mathscr{l}_{ce}(yf(\boldsymbol{W}^{(t)}, \tilde{\mathbf{x}}) \big]}
\end{equation}
where
\begin{equation}
    \mathscr{l}_{ce}(yf(\boldsymbol{W}^{(t)}, \tilde{\mathbf{x}}) \leq 2 + \frac{1}{m}\sum_{j=-y,r\in[m]}\sigma(\langle \mathbf{w}_{j,r}^{(t)}, \tilde{\boldsymbol{\xi}} \rangle)
\end{equation}

We can observe that the generalization gap of GNNs is positively correlated with the inner product of model parameters and noise. Next, we will prove why ensuring consistent norms for node representations within the same class can reduce the inner product of model parameters and noise, thereby reducing the generalization gap.

\begin{proof}
For the $k$-th class of samples, when the representation norms are consistent, we can derive that their covariance matrix is zero matrix $\boldsymbol{O}$, as shown below:
\begin{equation}
    \Sigma_{\boldsymbol{X}\boldsymbol{W}} = \frac{1}{n-1}(\boldsymbol{X}\boldsymbol{W}-\boldsymbol{M})(\boldsymbol{X}\boldsymbol{W}-\boldsymbol{M})^{\top} = \boldsymbol{O}
\end{equation}
where $\boldsymbol{M} \in \mathbb{R}^{n \times 2d}$ denotes the mean of the $k$-th class samples' representations.

Then, we have
\begin{align}
    (n-1)\Sigma_{\boldsymbol{X}\boldsymbol{W}} 
    &=  (\boldsymbol{X}\boldsymbol{W}-\boldsymbol{M})(\boldsymbol{X}\boldsymbol{W}-\boldsymbol{M})^{\top} \\
    &= ((\boldsymbol{\mu} + \boldsymbol{\xi})\boldsymbol{W} - (\boldsymbol{M}_{\boldsymbol{\mu}} + \boldsymbol{M}_{\boldsymbol{\xi}}))\\
    &\cdot((\boldsymbol{\mu} + \boldsymbol{\xi})\boldsymbol{W} - (\boldsymbol{M}_{\boldsymbol{\mu}} + \boldsymbol{M}_{\boldsymbol{\xi}}))^{\top}\\
    &=(\boldsymbol{\mu}\boldsymbol{W}-\boldsymbol{M}_{\mu})(\boldsymbol{\mu}\boldsymbol{W}-\boldsymbol{M}_{\mu})^{\top} \\
    &+ (\boldsymbol{\xi}\boldsymbol{W}-\boldsymbol{M}_{\xi})(\boldsymbol{\xi}\boldsymbol{W}-\boldsymbol{M}_{\xi})^{\top}\\
    & = (n-1) \Sigma_{\boldsymbol{\mu}\boldsymbol{W}} + (n-1) \Sigma_{\boldsymbol{\xi}\boldsymbol{W}}\\
    &= \boldsymbol{O}
\end{align}

Finally, we have
\begin{align}
\begin{cases}
\boldsymbol{\mu}\boldsymbol{W} = \boldsymbol{M}_{\mu}(\boldsymbol{\mu}\boldsymbol{W} \neq \boldsymbol{O}) \\
\boldsymbol{\mu}\boldsymbol{W} = \boldsymbol{O} & \text{which doesn't hold. }\\
\boldsymbol{\xi}\boldsymbol{W} = \boldsymbol{M}_{\xi}(\boldsymbol{\xi}\boldsymbol{W} \neq \boldsymbol{O}) & \text{which doesn't hold. } \\
\boldsymbol{\xi}\boldsymbol{W} = \boldsymbol{O} 
\end{cases}
\end{align}

We can see that as the representation norms become more consistent, the inner product between the model parameters and noise $\boldsymbol{\xi}\boldsymbol{W}$ approaches $\boldsymbol{O}$, thus reducing the generalization gap and improving the model's OOD generalization ability.
    
\end{proof}

\subsection{Proof of Proposition \ref{proposition-inter-class-balanced}}\label{Appendix-proposition-inter-class-balanced}

\begin{proof}
According to related work on neural collapse, the representations of training samples in $k$-classification tasks eventually form an equiangular tight frame projected onto a hyperplane $\phi \in \mathbb{R}^{k-1}$. On $\phi$, we seek the direction of $\tilde{\boldsymbol{e}}$, which maximizes the distribution spread, i.e., the direction along which the sum of absolute values of the projections is maximized. Since we are only interested in the direction $\tilde{\boldsymbol{e}}$, we set $\tilde{\boldsymbol{e}}$ as a unit vector and compute the inner product of the sample representations $\mathbf{z}$ and $\tilde{\boldsymbol{e}}$, as follows:
\begin{align}
    \tilde{\boldsymbol{e}} 
    &= \mathop{argmax}\limits_{\boldsymbol{e}}\frac{1}{n}\sum_{i=1}^n |\mathbf{z}_i \cdot \boldsymbol{e}|    \\
    &= \mathop{argmax}\limits_{\boldsymbol{e}}\frac{1}{n}\sum_{i=1}^n (\mathbf{z}_i \cdot \boldsymbol{e})^2  \\
    &=\mathop{argmax}\limits_{\boldsymbol{e}}\frac{1}{n} \lVert \boldsymbol{Z}^{\top}\boldsymbol{e}  \rVert_2^2
\end{align}

Consider symmetric matrix $\boldsymbol{Z}^{\top}\boldsymbol{Z} \in \mathbb{R}^{d \times d}$, let $\lambda_1 \geq \lambda_2 \geq \cdots \geq \lambda_d$ be its $d$ eigenvalues, and corresponding eigenvectors $\boldsymbol{\xi}_1, \boldsymbol{\xi}_2, \cdots, \boldsymbol{\xi}_d$.

Take an arbitrary unit vector \(\boldsymbol{e} = \sum_{i=1}^{d}\alpha_i \boldsymbol{\xi}_i \) on the hyperplane $\phi$, we have
\begin{equation}
    \lVert \boldsymbol{e} \rVert_2^2 = \langle  \boldsymbol{e}, \boldsymbol{e}  \rangle = \alpha_1^2 + \cdots + \alpha_d^2
\end{equation}
Then, we have
\begin{align}
    \langle \boldsymbol{e}, \boldsymbol{Z}^{\top}\boldsymbol{Z}\boldsymbol{e} \rangle 
    &= \langle \alpha_1 \boldsymbol{\xi}_1 + \cdots +\alpha_d\boldsymbol{\xi}_d,  \alpha_1 \boldsymbol{Z}^{\top}\boldsymbol{Z}\boldsymbol{\xi}_1 + \cdots +\alpha_d \boldsymbol{Z}^{\top}\boldsymbol{Z}\boldsymbol{\xi}_d \rangle\\
    &= \langle \alpha_1 \boldsymbol{\xi}_1+\cdots +\alpha_d\boldsymbol{\xi}_d,  \alpha_1 \lambda_1 \boldsymbol{\xi}_1 + \cdots +\alpha_d \lambda_d \boldsymbol{\xi}_d \rangle\\
    &= \lambda_1\alpha_1^2 + \lambda_2\alpha_2^2 + \cdots +\lambda_d\alpha_d^2\\
    &\leq \lambda_1(\alpha_1^2 + \alpha_2^2 + \cdots +\alpha_d^2)\\
    &= \lambda_1 \lVert \boldsymbol{e} \rVert^2_2\\
    &= \lambda_1
\end{align}

When the representation norms are consistent, the eigenvalues of the symmetric matrix $\boldsymbol{Z}^{\top}\boldsymbol{Z}$ satisfy:
\begin{align}
    \lambda_1 = \lambda_2 = \cdots = \lambda_{k-1} > 0\\
    \lambda_k = \lambda_{k+1} = \cdots = \lambda_{d} = 0
\end{align}
which means $\forall \boldsymbol{e} = \sum_{i=1}^d \alpha_i \boldsymbol{\xi}_i$ on hyperplane $\phi$, we have $\langle \boldsymbol{e}, \boldsymbol{Z}^{\top}\boldsymbol{Z}\boldsymbol{e} \rangle 
 = \lambda_1$.
 Therefore, by enforcing consistent representation norms, we identified a hyperplane $\phi$ where the sum of the absolute values of the node features' projections is maximized, leading to the most significant separation between classes, which is more conducive to classification.
    
\end{proof}

\begin{table}[!t]
\centering
\caption{Comparison of different methods that promote consistency in node representation norms under OOD generalization scenarios with GCN as the backbone. We combine these methods with BIM and compare the Accuracy(↑) and running times (s).}\label{tabel-LOSS-OOD}
\resizebox{\linewidth}{!}{
\begin{tabular}{c|cc|cc}
\hline
\hline
\multirow{2}*{Method} &\multicolumn{2}{c|}{Cora} &\multicolumn{2}{c}{Citeseer} \\
~ &Accuracy	&Time (s)	&Accuracy	&Time (s) 			\\
\hline
\hline
CaNet & 96.1 ± 1.4 &33.2 ± 0.7 &94.6 ± 1.9  &36.5 ± 0.7\\   
w/ $\mathcal{L}_{scl}$ & 96.5 ± 1.5 &43.6 ± 1.1 &95.7 ± 1.5 &52.7 ± 1.3\\   
w/ $\mathcal{L}_{center}$ & 97.3 ± 1.9 &2,009.4 ± 4.2 &95.9 ± 1.1 &2,074.9 ± 5.7\\   
w/ $\mathcal{L}_{bound}$ & 97.8 ± 2.1 &33.3 ± 0.7 &96.0 ± 1.3 &36.7 ± 0.8\\   
w/ $\mathcal{L}_{\mathrm{NodeReg}}$(ours) & 98.6 ± 0.2 &33.3 ± 0.7 &97.5 ± 0.2 &36.7 ± 0.7\\   
\hline
\hline
\end{tabular}
}
\end{table}

\section{More Experiments}

\subsection{Comparison with Different Methods for Norm Consistency in OOD Generalization Scenario}\label{appendx-subsection-comparison-OOD-loss}


We compare the accuracy and runtime of loss functions that impose varying degrees of consistency constraints on node representation norms in OOD (out-of-distribution) generalization scenarios. The results in Table \ref{tabel-LOSS-OOD} demonstrate that enforcing norm consistency within class node representations improves the accuracy of semi-supervised node classification using GCN in this setting. Furthermore, enforcing norm consistency between class node representations by maximizing inter-class distances further enhances the classification accuracy, indicating that our proposed method, NodeReg, achieves optimal performance.


\begin{figure}[!t]
	\centering
	\includegraphics[width=0.9\linewidth]{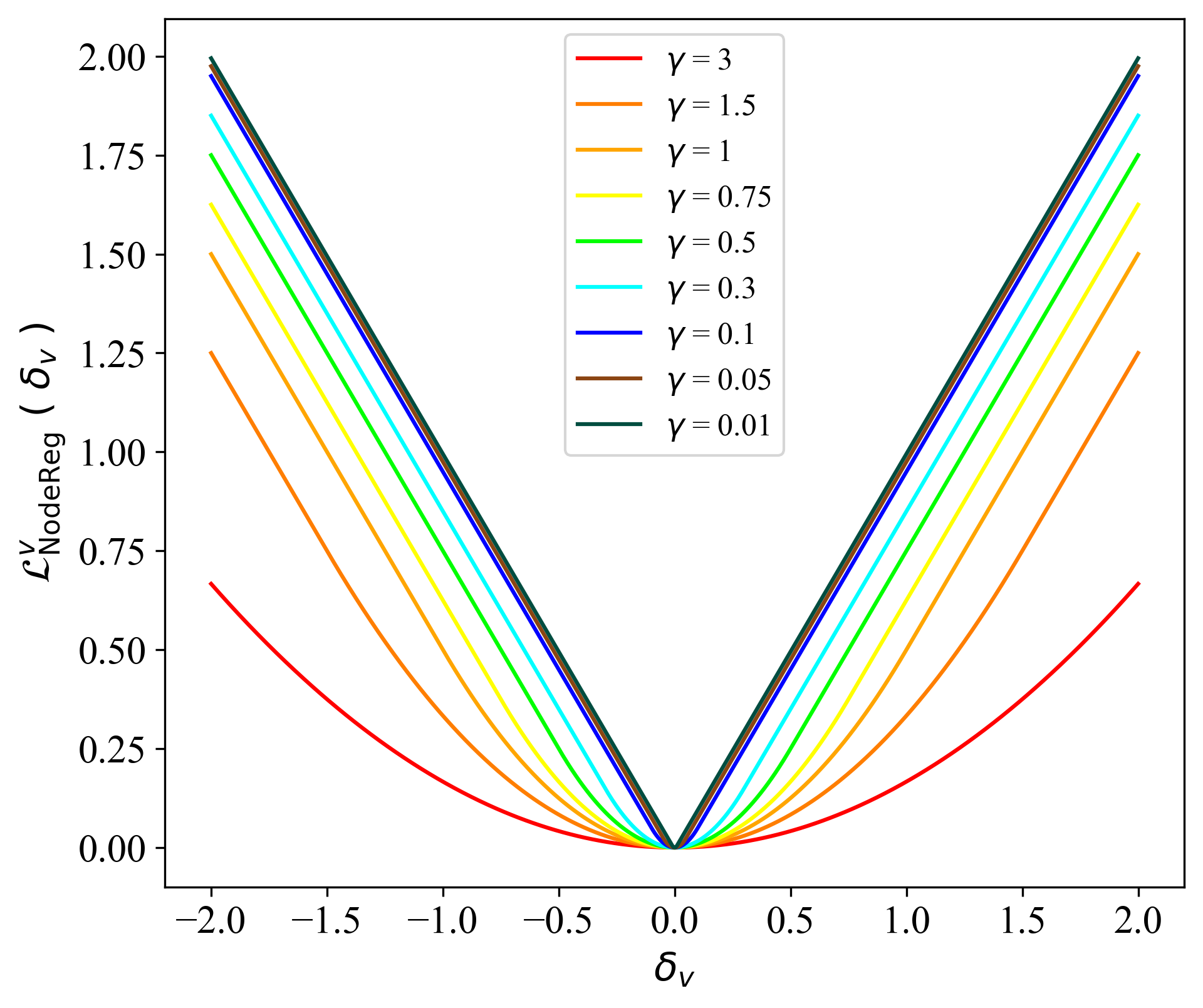}
  \Description{Gamma}
    \caption{$\mathcal{L}_{\mathrm{NodeReg}}^v(\delta_v)$}
    \label{F-gamma}
\end{figure}

\end{document}